\PassOptionsToPackage{prologue,dvipsnames,table}{xcolor}
\documentclass[sigconf,natbib=True]{acmart}

\AtBeginDocument{%
  }

\copyrightyear{2024}
\acmYear{2024}
\setcopyright{rightsretained}
\acmConference[SIGSPATIAL '24]{The 32nd ACM International Conference on
Advances in Geographic Information Systems}{October 29-November 1,
2024}{Atlanta, GA, USA}
\acmBooktitle{The 32nd ACM International Conference on Advances in
Geographic Information Systems (SIGSPATIAL '24), October 29-November 1,
2024, Atlanta, GA, USA}
\acmDOI{10.1145/3678717.3691256}
\acmISBN{979-8-4007-1107-7/24/10}

\usepackage{amsmath,amsfonts}
\usepackage{algorithmic}
\usepackage{algorithm}
\usepackage{array}
\usepackage[caption=false,font=normalsize,labelfont=sf,textfont=sf]{subfig}
\usepackage{textcomp}
\usepackage{stfloats}
\usepackage{url}
\usepackage{verbatim}
\usepackage{graphicx}
\usepackage{booktabs}
\usepackage{float}

\usepackage{amsmath}
\usepackage{mathtools}
\usepackage{amsthm}

\theoremstyle{plain}

\newtheorem{theorem}{Theorem}

\theoremstyle{definition}

\theoremstyle{remark}

\usepackage{listings}%

\definecolor{codepurple}{rgb}{0.58,0,0.82}
\definecolor{backcolour}{rgb}{0.95,0.95,0.92}

\lstdefinestyle{pytorchstyle}{
    backgroundcolor=\color{backcolour},
    commentstyle=\color{red},
    keywordstyle=\color{codepurple},
    stringstyle=\color{orange},
    basicstyle=\ttfamily\footnotesize,
    breakatwhitespace=false,
    breaklines=true,
    captionpos=b,
    keepspaces=true,
    showspaces=false,
    showstringspaces=false,
    showtabs=false,
    tabsize=2,
    language=Python,
    morekeywords={Tensor,Module,nn,Sequential,Linear,ReLU}
}

\begin{document}

\title{Enhancing Graph Neural Networks in Large-scale Traffic Incident Analysis with Concurrency Hypothesis }


\author{Xiwen Chen}
\author{Sayed Pedram Haeri Boroujeni}
\email{{xiwenc,shaerib}@g.clemson.edu}
\affiliation{%
  \institution{Clemson University}
  \city{Clemson}
  \state{SC}
  \country{USA}
}

\author{Xin Shu}
\email{shu.xin@northeastern.edu}
\affiliation{%
  \institution{Northeastern University}
  \city{Boston}
  \state{MA}
  \country{USA}
}

\author{Huayu Li}
\email{hl459@arizona.edu}
\affiliation{%
  \institution{University of Arizona}
  \city{Tucson}
  \state{AZ}
  \country{USA}
}

\author{Abolfazl Razi}
\email{arazi@clemson.edu}
\affiliation{%
  \institution{Clemson University}
  \city{Clemson}
  \state{SC}
  \country{USA}
}

%





\renewcommand{\shortauthors}{Xiwen et al.}

\begin{abstract}
 Despite recent progress in reducing road fatalities, the persistently high rate of traffic-related deaths highlights the necessity for improved safety interventions. Leveraging large-scale graph-based nationwide road network data across 49 states in the USA, our study first posits the Concurrency Hypothesis from intuitive observations, suggesting a significant likelihood of incidents occurring at neighboring nodes within the road network. To quantify this phenomenon, we introduce two novel metrics, Average Neighbor Crash Density (ANCD) and Average Neighbor Crash Continuity (ANCC), and subsequently employ them in statistical tests to validate the hypothesis rigorously. Building upon this foundation, we propose the Concurrency Prior (CP) method, a powerful approach designed to enhance the predictive capabilities of general Graph Neural Network (GNN) models in semi-supervised traffic incident prediction tasks. Our method allows GNNs to incorporate concurrent incident information, as mentioned in the hypothesis, via tokenization with negligible extra parameters. 
 The extensive experiments, utilizing real-world data across states and cities in the USA, demonstrate that integrating CP into 12 state-of-the-art GNN architectures leads to significant improvements, with gains ranging from 3\% to 13\% in F1 score and 1.3\% to 9\% in AUC metrics. The code is publicly available at \url{https://github.com/xiwenc1/Incident-GNN-CP}\footnote{We tend to use the term \texttt{incident} rather than \texttt{accident} according to the preference of the Department of Transportation. They may be interchangeably used in our paper.}.

\end{abstract}

\ccsdesc[500]{Information systems~Location based services}
\ccsdesc[500]{Information systems~Geographic information systems}
\ccsdesc[500]{Computing methodologies~Semi-supervised learning settings}

\keywords{Road Network Analysis, Graph Analysis, Graph Neural Network}


\maketitle

\begin{figure}[h]
    \centering
    \includegraphics[width=0.4\textwidth]{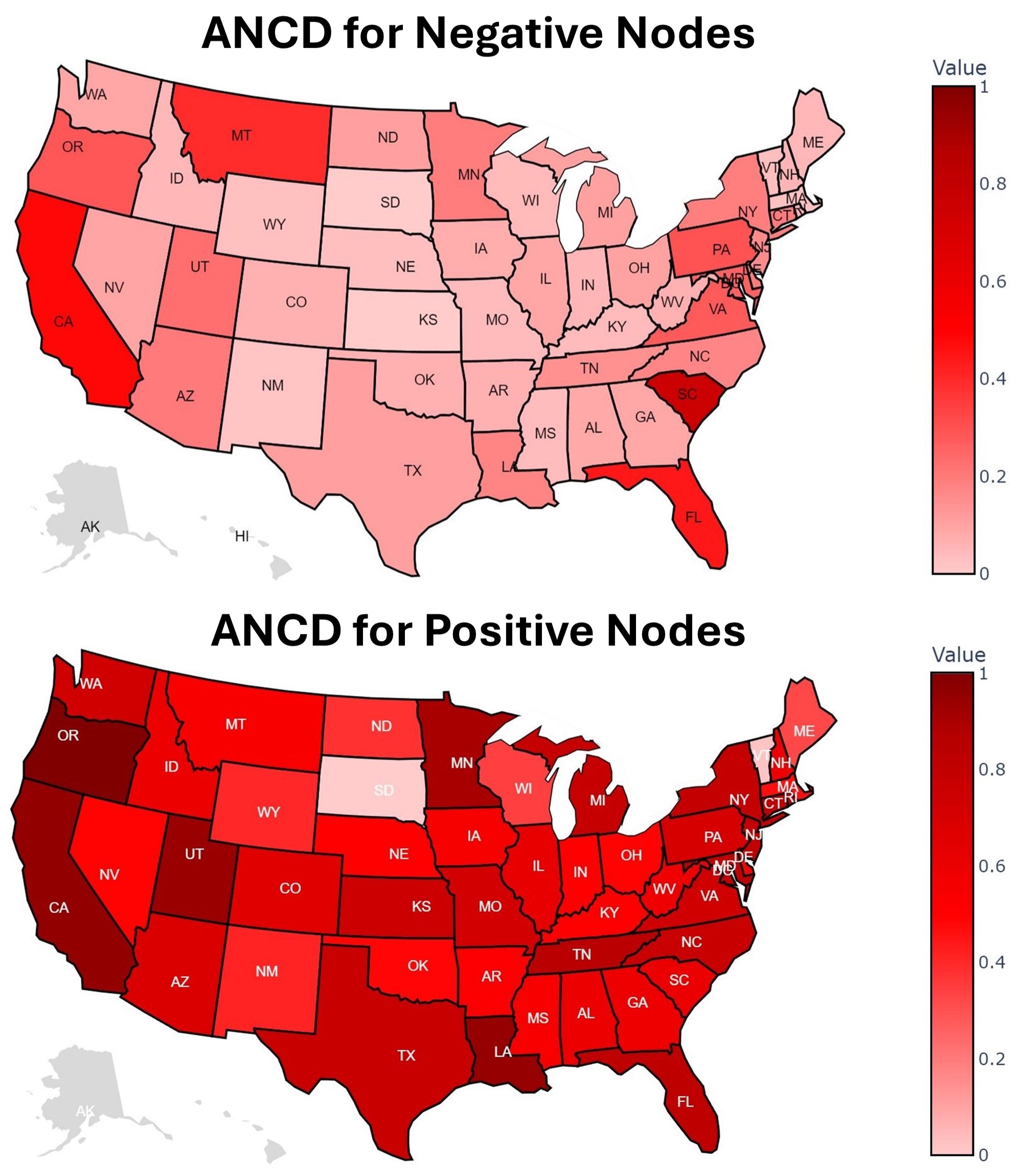}
    \vspace{-0.2cm}
    \caption{The state-wise Average Neighbor Crash Density (ANCD) for \textbf{Top:} negative nodes (i.e. nodes without incident records) and \textbf{Bottom:} positive nodes (i.e. nodes with incident records) when $k=1$.
    For a specific class of nodes (i.e. positive/negative nodes), a deeper color denotes a higher density of their neighbor nodes have incident. It is observed that, \underline{\textit{within the same state}}, the neighbor nodes of positive nodes often exhibit a higher crash density than those of negative, which supports our Concurrency Hypothesis. } 

    \vspace{-0.1cm}
    \label{fig:visuaul_ancd}
\end{figure}

\section{Introduction}

The significance of traffic safety is underscored by recent statistics, which highlight the ongoing challenges and successes in reducing road fatalities. In the early months of 2023, the National Highway Traffic Safety Administration (NHTSA) \cite{NHTSA} reported a decrease in traffic fatalities, estimating that 9,330 people lost their lives in traffic crashes in the first quarter, marking a 3.3\% decline compared to the previous year. This trend continued into the first half of the year, with fatalities dropping to 19,515, also down by about 3.3\% from the prior year. However, despite the positive trend in these numbers, they still reflect a high rate of traffic-related deaths, especially compared to the earlier years (2005-2019). This situation emphasizes the ongoing need for enhanced traffic safety measures and interventions to sustain and accelerate the reduction in road fatalities.

The importance of modeling traffic incident risks is well recognized in the field of urban planning and public safety \cite{fadhel2024comprehensive,razi2023deep}. Accurate predictions of where and when incidents are likely to occur can significantly contribute to the development of more effective traffic management strategies and infrastructure improvements. As urban areas continue to grow and also traffic volume increases, the need for powerful analytical tools to assess risk and prevent incidents becomes increasingly critical. It requires the integration of comprehensive datasets and advanced analytical techniques to understand the complex dynamics of road traffic and enhance safety measures. By leveraging detailed geospatial data and traffic incident records, researchers and city planners can identify high-risk areas and implement specific interventions to mitigate the possible risks, while enhancing road safety \cite{pei2024safety}. Recently, many studies have analyzed the effect of road features for predicting incident occurrences, such as \cite{persaud1992accident,oh2006regressionaccident,caliendo2007regressionaccident,najjar2017imgaccident,zhou2020riskoracle,zheng2021modeling}. 
More recently, 
Deep Learning (DL)-based methods have gained significant attention in traffic safety analysis since their powerful ability to characterize the inherent complex features of large-scale data \cite{yin2021deep,sarlak2023diversity,zhang2023multiclass,razi2023deep}.
Due to the nature that both the road network and traffic flow can be viewed as graph structure data, Graph Neural Network (GNN) \cite{kipf2016gcnconv,velivckovic2017gat,hamilton2017inductive,du2018tagcn,xie2018cgc,li2020gen,liu2022contrastive,zhang2024beyond} is the rational choice to characterize the relations in a network and has been adopted in recent works \cite{li2017speedrandomwalkgru,yu2021deep,zhou2020foresee}.




Our work is motivated by two intuitive observations in traffic incident occurrences in road networks: When people are driving and notice a traffic incident, there is a high probability that they observe another incident has occurred nearby. Another observation is that there are always some accident-prone sections, meaning the continuous areas included in the sections are likely to have incidents even if they have not occurred necessarily at the same time. We then make a unified hypothesis for them:

\textbf{Concurrency Hypothesis.} \textit{There is a high probability when a node has an incident occurred, some of its neighbors have an incident occurred.} 

Subsequently, we propose two novel metrics, the Average Neighbor Crash Density (ANCD) and Average Neighbor Crash Continuity (ANCC), to quantify these observations, and apply standard statistical tests for these quantitative results to validate the proposed hypothesis. An exemplary visualization of ANCD for each state is shown in Fig. \ref{fig:visuaul_ancd}, which underscores the difference of the nodes between different categories for each state. 
We then conjecture that this hypothesis indicates that there may be some important but difficult-to-capture information and features that have not been fully collected by the general datasets. 
Accordingly, in this work, we proposed an enhancement method called \textit{Concurrency Prior} (CP) that explores the hidden information beyond the common features from the crash label for semi-supervised traffic incident prediction. This problem is built on a single monolithic graph representing an entire state or city. Entire edge features, entire node features and partial nodes' labels are known. Our goal is to utilize the known information to learn a model and predict the label for the rest of the nodes with unknown labels. The formal problem description is in Section \ref{sec:pf}. Our method is compatible and complementary with general graph neural networks, such as Graph Convolutional Networks \cite{kipf2016gcnconv}, Graph Attention Networks \cite{velivckovic2017gat}, and Graph Transformers \cite{shi2021graphtransformer}. Our investigation is based on the nationwide real-world road network data provided by \cite{huang2023tap}. This large-scale data source contains the incident record from 49 states of the USA and provides various edge features, such as length, type, number of lanes, max speed, and road direction and angular information. We provide the details of the data acquisition in Section \ref{sec:data}. 

In summary, our contribution is two-fold: \textbf{(i)} We are the first to statistically validate the \textit{Concurrency Hypothesis} in nationwide graph-based data by using our proposed metrics; and \textbf{(ii)} We propose an enhancement method called \textit{Concurrency Prior} that enables boosting broad variations of graph neural networks in the semi-supervised traffic incident prediction task by introducing negligible parameters. 
Our intensive experiments on 12 state-of-the-art graph neural networks demonstrate a 3\%-13\% and 1.3\%-9\% gain in F1 and AUC, respectively.

\section{Related Work}








\begin{figure*}[h]
    \centering
    \includegraphics[width=0.7\textwidth]{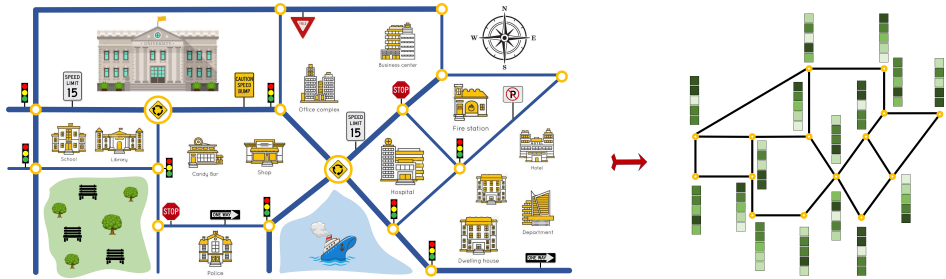}
    \vspace{-0.2cm}
    \caption{The graph-based data is obtained from real-world road networks.}

    \label{fig:dataprocess}
\end{figure*}

It is known that most traffic-related data can be viewed as network/graph structure data. For example, vehicles involved in an incident can be treated as a network, where each node is a vehicle, and the edge denotes the interaction between every two vehicles. Besides, the road network can be treated as a graph, where the nodes denote different physical locations, and the edges denote roads between them. 
Hence several previous works focus on network analysis \cite{chen2022network,wu2004urban,gao2007study,boccaletti2014structure,klinkhamer2017functionally} and many works analysis it based on knowledge from the complex network theories, such as small-world networks \cite{watts1998collective} and random scale-free
networks \cite{barabasi1999emergence}. However, this type of data is not as easy to process as usual data (e.g., images) in the machine learning community. This is due to the fact the topology of graph data is often variable and enormous \cite{prince2023understanding}. For example, considering a city as a graph, different cities apparently have totally different topologies and may have massive intersections as nodes. Therefore, graph Neural Networks (GNNs) have gained significant attention for processing graph data.

The early concept of GNNs can be traced back to 2008 \cite{scarselli2008graph}, when the authors proposed a framework that leverages a recurrent neural network (RNN) structure for graph data. Afterward, authors in \cite{bruna2013spectral,henaff2015deep} apply spectral approaches to GNNs, where they perform convolutions via graph Laplacian. However, spectral approaches introduce an intensive computation cost and lack generalizability across different graphs. To tackle these issues, authors in \cite{kipf2016gcnconv} propose Graph Convolutional Networks (GCNs), which significantly reduce computational complexity while maintaining performance. This method then becomes the cornerstone in the development of GNNs, and subsequently, there are a large number of variants \cite{velivckovic2017gat,hamilton2017inductive,du2018tagcn,xie2018cgc,li2020gen}. For example, 
authors in \cite{velivckovic2017gat} introduce attention mechanisms to model the importance of nodes' neighbors, while authors in \cite{hamilton2017inductive} propose an inductive framework that learns node embeddings by sampling and aggregating features from a node's local neighborhood. Authors in \cite{du2018tagcn} propose a topology adaptive mechanism for graph convolution. In 2020, authors in \cite{li2020gen} developed a framework for training very deep GCNs using differentiable generalized aggregation functions and a novel normalization layer called MsgNorm, effectively addressing vanishing gradients, over-smoothing, and over-fitting issues in very deep GCN models.
More recently, authors in \cite{shi2021graphtransformer} propose an adaptation of the transformer architecture to graph-structured data, providing an alternative to convolution-based methods.

GNNs illustrate the obvious superiority of capturing the dependencies of graph-based data in non-euclidean space, while these dependencies are challenging to learn by the algorithms designed for Euclidean space (e.g,. Convolutional Neural Networks (CNN)). 
Therefore, GNNs have been widely used in traffic and intelligence transportation fields \cite{ye2020build,rahmani2023graph}.
For instance, authors in \cite{zhao2019t,guo2019attention,shin2020incorporating,jiang2022graph} employ GNN for traffic flow prediction. Demand Prediction is also a popular task that can benefit from GNN. These demands include ride-hailing demand forecasting
 \cite{geng2019spatiotemporal,ke2021predicting,huang2022gan}, bike sharing systems \cite{lin2018predicting,li2022data}, and passenger flow prediction \cite{liu2020physical}. Additionally, GNN is used in point-cloud-based perception \cite{shi2020point,jing2022agnet}, motion prediction \cite{tang2023trajectory}, and planning \cite{cai2022dq} in the studies of autonomous vehicles. Likewise, there are several works aiming to predict the incident occurrence. Specifically, \cite{zhou2020foresee} develops a novel differential time-varying GCN to dynamically capture traffic variations and \cite{yu2021deep} proposes a spatio-temporal GCN and employs the embedding layer to remove noises and better extract semantic representations of external information. 

The most related work is \cite{huang2023tap}, which performs the traffic analysis with nationwide coverage and real-world network topology and tries to solve the classification problem solely based on a single monolithic graph. It is noteworthy that none of the previous works mentioned above has performed analysis on such a large data scale. Our work substantially enhances the prediction performance over this work, as well as several popular aforementioned GNN variants, by imposing our proposed concurrency prior to neural networks. The proposed training strategy mentioned in Section \ref{sec:trainnetwork} is related to attribute masking used in \cite{you2020graph,liu2022contrastive}; however, we use it to mimic the real node-wise inference in the training phase when incorporating the concurrency information (Eq. \ref{eq:loss}). Hence, our training strategy is essentially different and orthogonal from theirs, since they only employ it as a common data augmentation method.


\section{Data Acquisition}\label{sec:data}

In our study, we use nationwide traffic incident data consists over 1,000 U.S. city-level datasets and 49 U.S. state-level datasets \cite{huang2023tap}. 
In this section, we delve into the key concept behind the creation of graph-based traffic incident benchmarks with datasets that contain real-world geospatial features.
The traffic incident processing repository is developed by collecting a comprehensive set of raw data on traffic incidents \cite{US_accident}. It involves detailed information about incident records, the geographical layout of streets, and the relational structure of these locations represented in graph form. To enhance the utility of the incident location data, a reverse geocoding process is employed to convert geographic coordinates into more accessible address formats. Afterward, the crash information is integrated with the graph-structured data and geographical attributes to create a cohesive and structured dataset, as shown in Fig. \ref{fig:dataprocess}. The foundational data for these datasets are sourced from Microsoft Bing Map Traffic \cite{Bing}, extracted explicitly from the US-Accidents benchmarks. These datasets serve as a valuable source of information, documenting around 2.8 million traffic-related incidents over a period from January 2016 through December 2021. They provide a detailed account of traffic events during this time frame, offering insights into patterns and trends.

In these datasets, OpenStreetMap (OSM) \cite{boeing2017osmnx} is employed as the primary resource for obtaining geospatial data information. The collected data from OSM are enriched with a variety of geographical information, including roads, trails, railway stations, land Use, land cover, transport networks and natural landmarks like forests and rivers. This information is tagged under different OSM classes, which serve to present the specific characteristics of the geographic elements in the database, such as nodes (defining points like intersections), ways (paths or open areas), and edges (logical or physical relationships between elements). Hence, the datasets now have rich features, including but not limited to the type of road, the length of a road, the number of lanes, one-way indication, max speed, tunnel indication, junction type, etc. Besides, directional and angular features of road networks are identified, enhancing the dataset with unique geometric insights.


Afterward, incident data is first reverse geocoded to pinpoint exact locations, and then systematically organized based on settlement hierarchies from villages to states. Datasets are divided into two main subsets: city-level and state-state, each tailored to different scales of traffic analysis. The city-level datasets focus on urban areas where traffic incident frequency is higher, reflecting the denser road networks and population distribution. In contrast, state-level datasets provide a broader perspective, suitable for regional traffic trends and policy planning. Eventually, the integration process involves sophisticated data processing techniques like one-hot encoding and spatial analysis used to correlate accident sites with nearby road network nodes.  

\begin{figure}[htbp]
    \centering
    \includegraphics[width=0.25\textwidth]{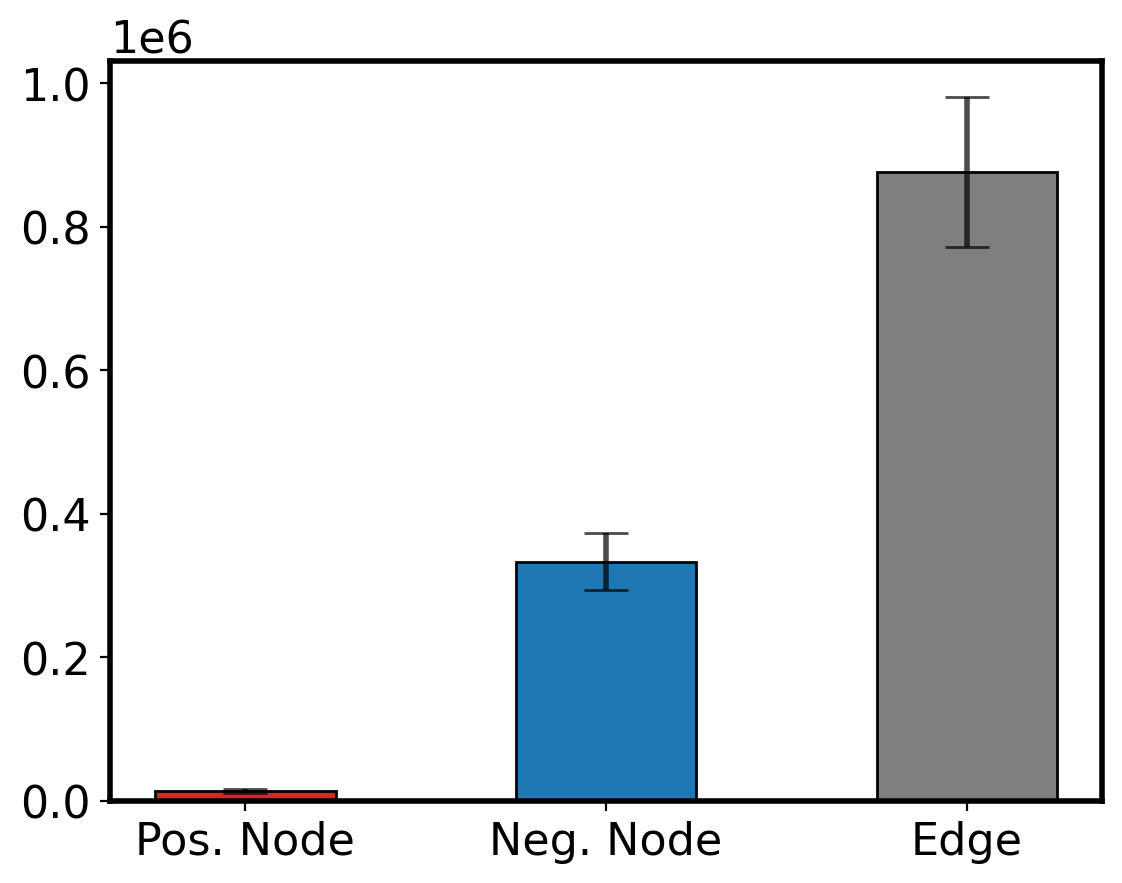}
    \vspace{-0.2cm}
    \caption{The statistics of the graph-based datasets on 49 states. }

    \label{fig:datastat}
\end{figure}

In summary, each dataset (either a city or a state) is a single monolithic graph, which refers to a unified and comprehensive graph structure that includes all data points (nodes) and relationships (edges) contained within one comprehensive graph without division into subgraphs. Suppose a dataset (can be a specific state or city) is a large graph that has $N$ nodes and $E$ edges, then the dataset can be presented by three matrices $\boldsymbol{A}\in\mathbb{R}^{N\times N}$, $\boldsymbol{X}\in\mathbb{R}^{ N\times D_1}$, and $\boldsymbol{E}\in\mathbb{R}^{ E\times D_2}$, and $\boldsymbol{Y}\in\mathbb{R}^{ N}$, denoting the adjacency matrix, node embedding, edge embedding, and node labels, respectively. Here, $D_1$ and $D_2$ denote the number of dimensions of node and edge features, respectively. It also should be noted that these datasets are significantly unbalanced, and a very low ratio of points is positive (i.e. nodes with crash records). The statistics of the node labels are shown in Fig. \ref{fig:datastat} and Table \ref{tab:datasets}. We realize this may substantially challenge most machine learning algorithms. 


\section{Problem Formulation}\label{sec:pf}

\begin{figure*}[!t]
    \centering
    \includegraphics[width=0.8\textwidth]{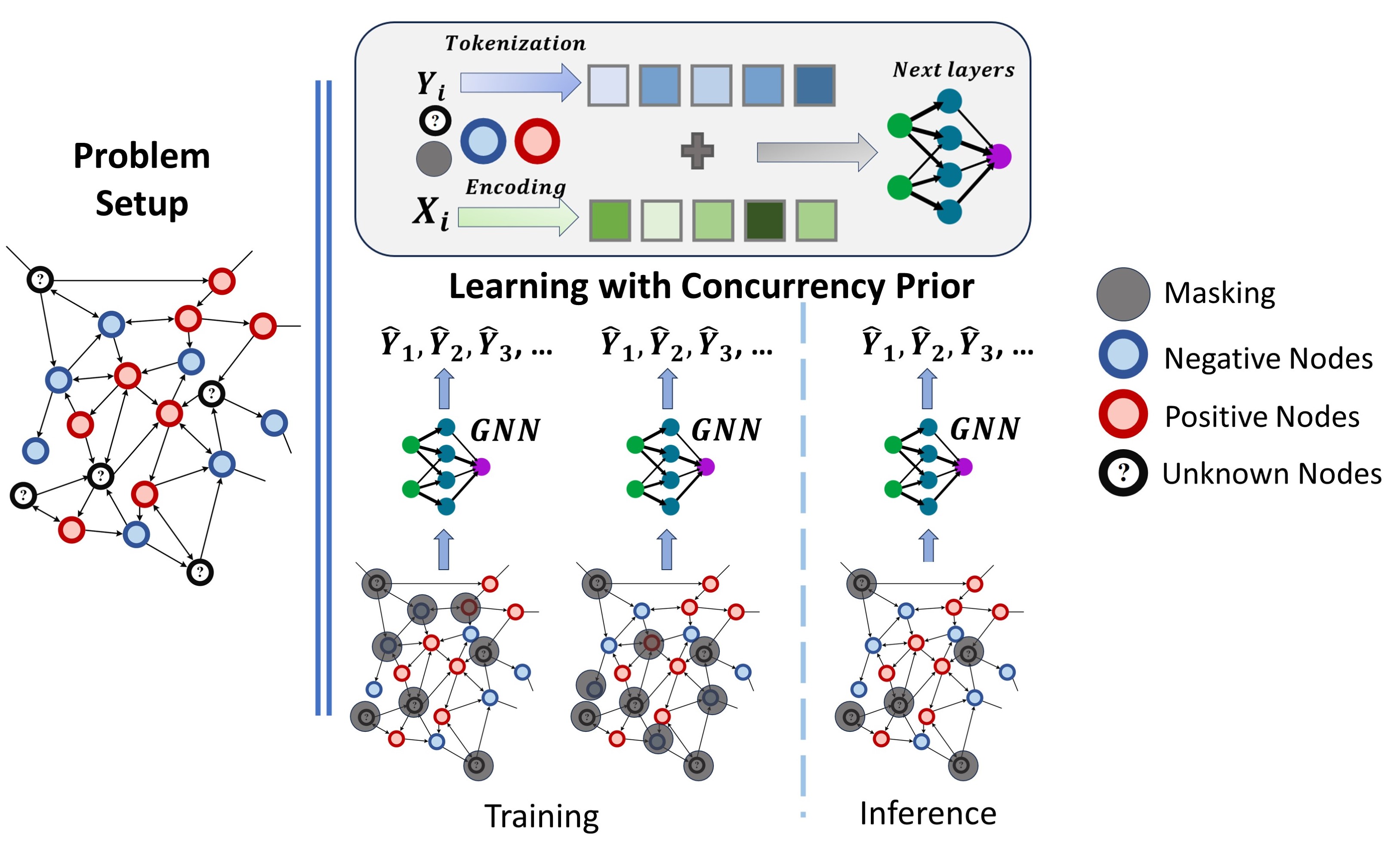}
    \vspace{-0.2cm}
    \caption{\textbf{Left:} The problem is formulated in a single large graph. Nodes' labels are known if the node is marked in colors, i.e., red (positive)/ blue (negative). The nodes with question marks are expected to be predicted. \textbf{Middle:} In the training phase, we keep all unknown nodes with the uncertain token 'o', and in each iteration, we also randomly mask some nodes with known labels to 'o' to mimic the prediction process. \textbf{Right:} In the inference phase, we only keep to nodes to be predicted with the uncertain token 'o'. \textbf{Top:} Imposing concurrency prior to the neural network.  }

    \label{fig:main1}
\end{figure*}

Here, we give the problem formulation of our study.
Suppose the label for a node $i$ is $\boldsymbol{Y}_i$. Our problem is semi-supervised and in a transductive setting. It means some nodes have known labels ($\boldsymbol{Y}_i,\forall i\in \mathcal{V}_{train}$), and others are unknown ($\boldsymbol{Y}_i,\forall i\in \mathcal{V}_{test}$) that we aim to predict them. Note that $\boldsymbol{A}$, $\boldsymbol{X}$, and $\boldsymbol{E}$ are fully known in both the training and inference phases.
We use $\boldsymbol{Y}_i=1$ to denote the positive nodes that there is an incident occurred while $\boldsymbol{Y}_i=0$ denotes the negative nodes that nothing happened here.
An illustration of the problem is shown in Fig. \ref{fig:main1} (\textbf{Left}), where red and blue denote the nodes with known labels (crash/no crash), and the question marks denote the unknown labels.

\section{Methodology}
In this section, we first adopt statistical tools to validate the concurrency hypothesis (Section \ref{sec:stat}), and then motivated by the effectiveness of this hypothesis, we propose the concurrency prior, which is an enhancement method for general graph neural networks in crash prediction (Section \ref{sec:proposed} and Section \ref{sec:trainnetwork}).

\subsection{Statistical Analysis of the Concurrency Hypothesis }\label{sec:stat}
Recap our \textbf{Concurrency Hypothesis}: \textit{There is a high probability when a node has an incident occurred, some of its neighbors have an incident occurred.} 

To validate this hypothesis, we propose two quantitative tools, and we expect to demonstrate that there is a statistical difference between the incident occurrence of the neighbors of positive nodes and negative nodes. The metrics are calculated state by state as each state is a monolithic graph.

\noindent\textbf{Average Neighbor Crash Density (ANCD).} This metric is first calculated for each node of a dataset (i.e. a state here) as,
\begin{align}
  NCD_i = \frac{\sum_{j\in neighbor_k(i)}\mathbb{I}(\boldsymbol{Y}_j=1)}{|neighbor_k(i)|}.
\end{align}
where $neighbor_k(i)$ denotes the set of neighbor nodes of node $i$ that can arrive in most $k$ hops through the connected edges. $\mathbb{I}(\cdot)$ and $|\cdot|$ denote the indicator function and the cardinality of a set, respectively.
Then, ANCD can be computed for positive nodes ($z=1$) and negative nodes ($z=0$) of the dataset, respectively, 
\begin{align}\label{eq:ANR}
  ANCD_z = \frac{\sum_{i\in \{ i| {Y}_i=z  \}}  NCD_i }{|\{ i| {Y}_i=z  \} |},
\end{align}
where $z\in \{0,1\}$.
 ANCD can be interpreted as the average density of the neighbor nodes that have crashes for a specific class of nodes.

We are also interested in the distance of the nearest positive nodes of nodes in a specific class. However, in a super-large graph, computing the distance in the form of hops is challenging due to the computation cost and memory issues; therefore, we propose the surrogate metrics to estimate it.

\noindent\textbf{Average Neighbor Crash Continuity (ANCC).} The metric aims to calculate how the continuity of neighbor crash nodes. 
\begin{align}
&  NCC_i = 
\left\{
\begin{aligned}
 &1,\quad \exists\boldsymbol{Y}_j=1,j\in neighbor_k(i) ,\\
&0,\quad \text{Else}.
\end{aligned}
\right.
\end{align} 
The $NCC_i$ can be interpreted as follows: if $NCC_i$ is equal to 1, the nearest positive node of node $i$ is at most $k$ hops. In contrast, if $NCC_i$ is equal to 0, the nearest positive node of node $i$ requires at least $k+1$ hops.
Subsequently, we can compute $ANCC_z$ similar to Eq. \ref{eq:ANR},
\begin{align}
  ANCC_z = \frac{\sum_{i\in \{ i| {Y}_i=z  \}}  HR_i }{|\{ i| {Y}_i=z  \} |}.
\end{align}
 Hence $ANCC_z$ is able to evaluate the average distance to the nearest positive nodes of the class $z$ nodes from a different perspective.



To comprehensively evaluate the difference between negative nodes and positive nodes, after computing the two metrics for each state, \textit{paired t-test} is used to offer support from the standard hypothesis test. 
To perform a paired t-test, we first compute the difference between paired observations, denoted as $d^j = M^j_0 - M^j_1$, where $M$ denotes one of the proposed metrics with predefined $k$, and $j$ denotes the state index.
Suppose the $\mu_d$ represents the population mean difference. The null hypothesis ($H_0$) and alternative hypothesis ($H_a$) for the paired t-test are typically defined as follows:
\begin{itemize}
    \item Null hypothesis ($H_0$): There is no significant difference between the paired observations, i.e., $\mu_d = 0$.
    \item Alternative hypothesis ($H_a$): There is a significant difference between the paired observations, i.e., $\mu_d < 0$.
\end{itemize}
If there is statistical significance, the incident occurrence in a node's neighbors that has a correlation with the status of this node. We present the results in Section \ref{sec:StatisticalResults}.




    


\subsection{Graph Neural Networks with Concurrency Prior}\label{sec:proposed}
As the Concurrency Hypothesis exists, we conjecture that there is information included in the label $\boldsymbol{Y}_i$ in addition to the original features $\boldsymbol{X}$ and $\boldsymbol{E}$. Hence, employing this information may enhance the model's learning ability. Here, a prediction for a node $i$ by a conventional graph neural work $G$ in the problem is presented as,
\begin{align}
    \hat{\boldsymbol{Y}}_i = G_i(\boldsymbol{A}, \boldsymbol{X}, \boldsymbol{E}).
\end{align}

In our method, we want to explicitly adopt the information, which results in a prediction as,
\begin{align}
    \hat{\boldsymbol{Y}}_i = F_i(\boldsymbol{A}, \boldsymbol{X}, \boldsymbol{E}, \{\boldsymbol{Y}_j| j\in\mathcal{V}_{train}\} ).
\end{align}
\begin{theorem}
    If we use the mutual information $I(\cdot;\cdot)$ to denote the upper bound of the learning ability of a network, apparently, $F_i$ should have a stronger potential of learning ability. This is because,
\begin{align}
    I(\boldsymbol{Y}_i; \boldsymbol{A}, \boldsymbol{X}, \boldsymbol{E} )\leq I(\boldsymbol{Y}_i; \boldsymbol{A}, \boldsymbol{X}, \boldsymbol{E},\{\boldsymbol{Y}_j| j\in\mathcal{V}_{train}\}  ).
\end{align}
\end{theorem}

Since the concurrency information is usually presented as discrete labels, it is now impossible to directly present any semantic information to the neural network. Therefore, we tokenize labels as
a learnable dictionary, and each instance (one vector) of the dictionary represents the latent feature of each category. This strategy is much more friendly for learning the neural network because, with tokenization, all operations are in continuous space, which allows us to optimize a neural network with concurrency information just like training a common network. Additionally, we use an efficient way to embed the concurrency information without introducing considerable parameters. Thereby, the \textit{Concurrency Prior} can be imposed to the neural network as,
\begin{align}
    \boldsymbol{X}_i\leftarrow encode(\boldsymbol{X}_i)+Tokenization (\boldsymbol{Y}_i).
\end{align}
We impose Concurrency Prior in the embedding space because the original features consist of data from different concepts (see Section \ref{sec:data}), and the $encode(\boldsymbol{X}_i)$ can be viewed as these feature after fusion, which offers a better representation. We aggregate the feature information ($\boldsymbol{X}_i$) and Concurrency Prior ($Tokenization (\boldsymbol{Y}_i)$) by summation because this way allows us not to change the original network architecture and hence not introduce extra parameters (except the few parameters by tokenization). For example, if the original network has one linear layer with $d_1\times d_2$ parameters. Alone with the parameters introduced by tokenization, if aggregating by concatenating, the architecture should be modified and has  $((d_1+d_{cp})\times d_2$ parameters accordingly, where $d_1$,$d_2$, and $d_{cp}$ are the number of the original input, output, and CP dimensions. In contrast, the total parameters in our method are consistently $d_1\times d_2$.
Hence, the total number of introduced parameters by imposing Concurrency Prior is $(C+1)\times d$ for the set of learnable vectors for tokenization, where $C$ and $d$ denote the number of classes and the embedding size of the feature in the original architecture, respectively. The additional one (i.e. $1$ in $C+1$) in the term denotes the uncertain class, which we will discuss in the next section.
When $C=2$ in our case, the introduced parameters are negligible.

\subsection{How to train the neural network?}\label{sec:trainnetwork}
To train the neural network, for each node $i\in\mathcal{V}_{train}$, we anticipate minimizing the loss for each node,
\begin{align}\label{eq:loss}
    \min_{F_i} \mathcal{L}( {\boldsymbol{Y}}_i,F_i(\boldsymbol{A}, \boldsymbol{X}, \boldsymbol{E}, \{\boldsymbol{Y}_j| j\in\mathcal{V}_{train}\setminus \boldsymbol{Y}_i \} )),
\end{align}
where $F_i$ denotes the classifier for node $i$.
Another challenge is posed here since a general graph neural network is designed to process a graph with arbitrary shapes, and training a classifier for each node is inefficient due to the massive number of nodes (e.g., 1169400 nodes in California dataset); therefore, the network often has a unified classifier for all nodes.
A general training strategy is feeding $\boldsymbol{A}, \boldsymbol{X}, \boldsymbol{E}$ to the network to predict all $\boldsymbol{Y}_i$, which assumes the prediction for all nodes uses the same input (i.e. $\boldsymbol{A}, \boldsymbol{X}, \boldsymbol{E}$), where our network is not fulfilled. In our case, 
we need to feed the feature and label of a node to the network; however, the label of the nodes from the test set is unknown, and we also need to exclude the label information of the target node (i.e. $\hat{\boldsymbol{Y}}_i=F_i(\boldsymbol{A}, \boldsymbol{X}, \boldsymbol{E}, \{\boldsymbol{Y}_j| j\in\mathcal{V}_{train}\setminus \boldsymbol{Y}_i \} )$) during training. To tackle these issues, we first introduce the uncertain token \textit{o} as a placeholder for the nodes without knowing the label information (i.e. test set). Then, in each iteration, we randomly set the labels of partial training nodes to \textit{o} to mimic the prediction processing that excludes the label information of the target nodes (shown in Fig. \ref{fig:main1} (\textbf{Middle})). With these proposed methods, we can train any graph neural network with Concurrency Prior in the same way as a common network. During inference, we will feed all known labels of training nodes to the GNN for the prediction (shown in Fig. \ref{fig:main1} (\textbf{Right})).  A summary of our proposed method is presented in Algorithm 1.

\begin{algorithm}
\begin{lstlisting}[style=pytorchstyle]
#input:  Hidden dim: d, number of classes: C, Node feature: X (N*D1), Adj. Matrix: A  (N*N), Edge Feature: E  (N*D2), hard label: Y (shape N*1), and train/valid/test indices: V_train, V_valid, V_test. Y[V_test] is unknown. Mask rate: R (0<R<1).

#output: Predicted probability for each node.
#The loss is only computed for all training nodes.


class GNNwithCP(torch.nn.Module):
    def __init__(self, hidden_dim=d,number_class =C):
        super(GNNwithCP, self).__init__()
        #define CP embedding
        self.CP_embeeding = nn.Embedding(dataset.num_classes+1,hidden_dim) #one class for masked nodes
        
        
        self.encoder_1 = ...  #output size should be hidden_dim
        self.encoder_other = ...
        self.fc = nn.Linear(..., C)#The unified classifier for all nodes 

    def forward(self,X,A,E,Y,M):
        
        token = torch.zeros_like(Y).to(Y.device)
        token[V_train] =  Y[V_train]+1 # We only know training nodes' labels, and others set to 0 meaning unknown. Original label 0->1, 1->2.
        
        if self.train: #only masking during training.
            select_index = random.sample(range(len(Y)),int(R*len(Y)))
            token[select_index]=0
        
        token_embeeding = self.CP_embeeding(token)
        X,E = self.encoder_1(X,A,E)
        X = X+token_embeeding #Eq. 8
        
        X,E = self.encoder_other(X,A,E)
        X = self.fc(X)
        return F.log_softmax(X, dim=1)

\end{lstlisting}
\caption{PyTorch Code for a general GNN with CP.}
\label{alg:penalty}
\end{algorithm}

\section{Experiment}

\subsection{Statistical Analysis Results}\label{sec:StatisticalResults}
\begin{figure}[]
    \centering
    \includegraphics[width=0.44\textwidth]{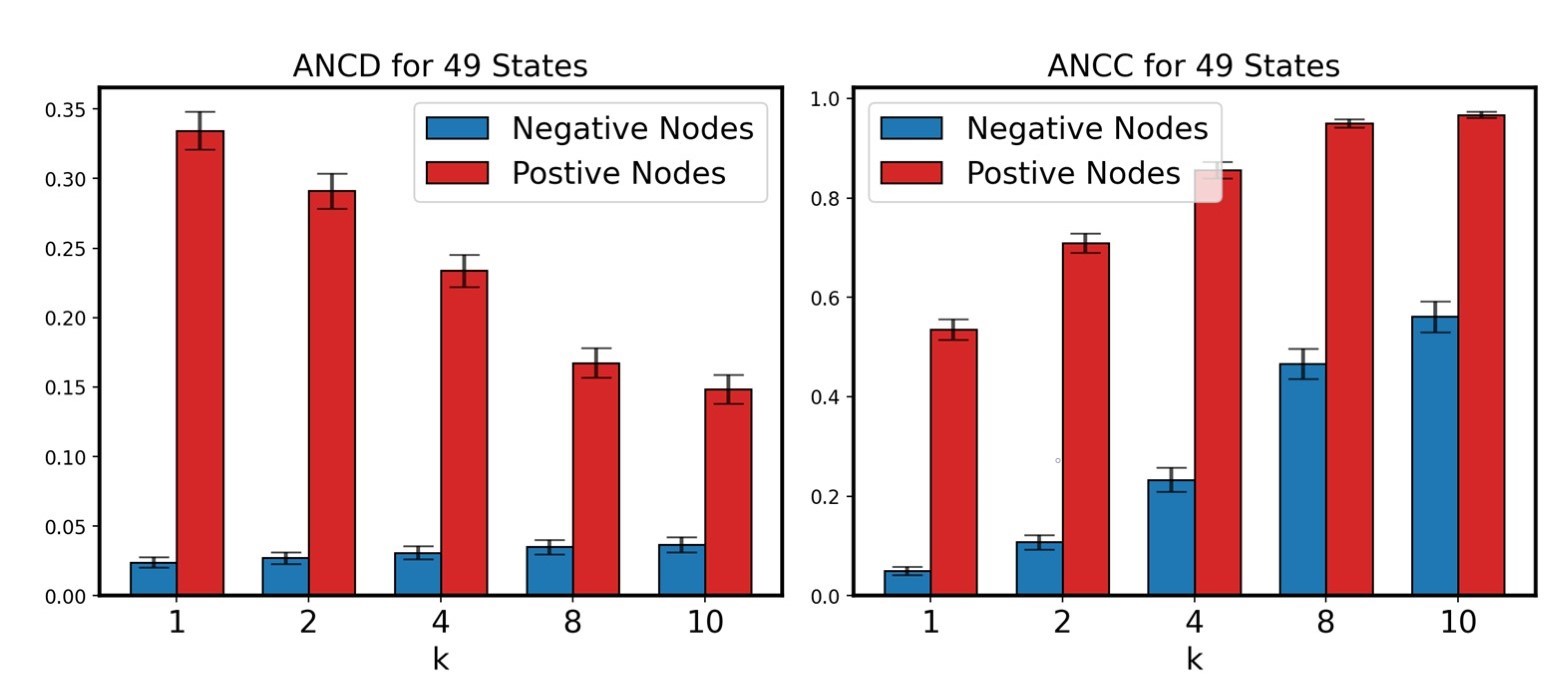}
    \vspace{-0.2cm}
    \caption{The statistics of ANCD and ANCC computed for the available 49 states.}

    \vspace{-0.1cm}
    \label{fig:metrics_value}
\end{figure}

We consider the number of available hops $k$ to $k\in\{1,2,4,8,10\}$ in our experiments.
The results of these paired t-tests conducted in Section \ref{sec:stat} are shown in
Table \ref{tab:pvalue}, where p-values are tiny for all tests (i.e. less than 1E-18). These results exhibit that we have very high confidence to conclude: in each state, the metrics ANCD and ANCC computed for positive nodes are statically higher than those for negative nodes, which supports our hypothesis that if a node has an incident occurred, its neighbors are likely to have incidents. A summary of the metrics is presented in Fig. \ref{fig:metrics_value}, which illustrates another interesting observation. We find that as the $k$ increases, the difference between the value of positive nodes and negative nodes decreases, which may suggest that the concurrency hypothesis has a high locality that indicates a label of a node is related to its closer neighbors. We also present the metrics for each state in Fig. \ref{fig:visuaul_ancd} to highlight their difference. 

\begin{table}[htbp]
\centering

\caption{The p-value of the paired-test. $k$ denotes the number of available hops.}
\vspace{-0.2cm}
\label{tab:pvalue}
\resizebox{0.45\textwidth}{!}{%
\begin{tabular}{cccccc}\\ \toprule
\textbf{k} & \textbf{1} & \textbf{2} & \textbf{4} & \textbf{8} & \textbf{10} \\ \midrule
\textbf{ANCD} & 1.13E-29 & 1.90E-28 & 5.84E-26 & 2.69E-21 & 4.53E-19 \\
\textbf{ANCC} & 6.20E-32 & 2.13E-37 & 3.42E-34 & 1.56E-23 & 1.05E-19 \\ \bottomrule
\end{tabular}%
}
\end{table}

\subsection{Main Experiment}

\noindent\textbf{Datasets.} We evaluate our proposed methods both city-wise and state-wise. Following \cite{huang2023tap}, we choose six representative cities, including Miami, Los Angeles, Orlando, Dallas, Houston, and New York. We also choose six representative states, including California, Oregon, Utah, Maryland, Minnesota, and Connecticut. A summary of these datasets is presented in Table \ref{tab:datasets}. All datasets
are stratified split into 60\% training/20\% validation/20\% testing. 

\noindent\textbf{Baselines.} We select 12 state-of-the-art GNN models, including (1) \textbf{GCN}: Graph Convolutional Networks \cite{kipf2016gcnconv}, (2) \textbf{ChebNet}: Chebyshev spectral graph convolution networks \cite{defferrard2016chebnet}, (3) \textbf{ARMANet}: Graph neural networks with convolutional auto-regressive moving average (ARMA) filters \cite{bianchi2021armanet}. (4) \textbf{GraphSAGE}: A general framework for inductive representation learning on graphs \cite{hamilton2017inductive}, (5) \textbf{TAGCN}: Topology adaptive graph convolutional networks \cite{du2018tagcn}, (6) 
\textbf{GIN}: Graph Isomorphism Networks \cite{xu2018gnn,weisfeiler1968reduction}, (7) \textbf{GAT}: Graph attention networks \cite{velivckovic2017gat}, (8) \textbf{MPNN}: Message Passing Neural Network \cite{gilmer2017mpnn}, (9) \textbf{CGC}: Crystal graph convolutional neural network \cite{xie2018cgc}, (10) \textbf{GEN}: GENeralized graph convolutional neural networks \cite{li2020gen}, (11) \textbf{Graphformer}: Graph transformers \cite{shi2021graphtransformer}, and (12) \textbf{TRAVEL}: a GNN designed for road network analysis \cite{huang2023tap}.

\noindent\textbf{Evaluation Metrics.} We use F1 score and Area Under the Receiver Operating Characteristic Curve (AUC) to evaluate the performance, since these datasets are obviously unbalanced, as presented in Section \ref{sec:data} as well as Table \ref{tab:datasets}.

\noindent\textbf{Implementation Detail.} Our training implementation strictly follows \cite{huang2023tap} and keeps the exact same structure and training hyperparameters (e.g., Optimizer, learning rate, weight decay, dropout, etc.). More details can be found in our source code. We find the reported results in \cite{huang2023tap} are highly reproducible; hence, we directly use their results in our comparison. For our methods, we report the average result and its standard deviation over 10 runs.


\begin{table}[]
\centering
\caption{The description of the selected datasets in our experiments.}
\vspace{-0.2cm}
\label{tab:datasets}
\resizebox{0.48\textwidth}{!}{%
\begin{tabular}{c|cccccc}\toprule
\textbf{Dataset} & \textbf{California} & \textbf{Oregon} & \textbf{Utah} & \textbf{Maryland} & \textbf{Minnesota} & \textbf{Connecticut} \\ \midrule
\textbf{\# of Nodes} & 1169400 & 217619 & 142478 & 234475 & 370383 & 120642 \\
\textbf{\# of Edges} & 2915853 & 544776 & 362667 & 557249 & 965962 & 304417 \\
\textbf{positive Ratio} & 0.106 & 0.072 & 0.059 & 0.057 & 0.052 & 0.042 \\ \midrule
\textbf{Dataset} & \textbf{\begin{tabular}[c]{@{}c@{}}Miami \\ (Florida)\end{tabular}} & \textbf{\begin{tabular}[c]{@{}c@{}}Los Angeles\\  (California)\end{tabular}} & \textbf{\begin{tabular}[c]{@{}c@{}}Orlando \\ (Florida)\end{tabular}} & \textbf{\begin{tabular}[c]{@{}c@{}}Dallas\\ (Texas)\end{tabular}} & \textbf{\begin{tabular}[c]{@{}c@{}}Houston\\ (Texas)\end{tabular}} & \textbf{\begin{tabular}[c]{@{}c@{}}New York\\ (New York)\end{tabular}} \\ \midrule
\textbf{\# of Nodes} & 8461 & 49251 & 7513 & 36150 & 59711 & 55404 \\
\textbf{\# of Edges} & 22648 & 135547 & 18216 & 92348 & 148937 & 140005 \\
\textbf{positive Ratio} & 0.133 & 0.130 & 0.302 & 0.258 & 0.221 & 0.083 \\ \bottomrule
\end{tabular}%
}
\end{table}

\begin{figure*}[]
    \centering
    \includegraphics[width=0.7\textwidth]{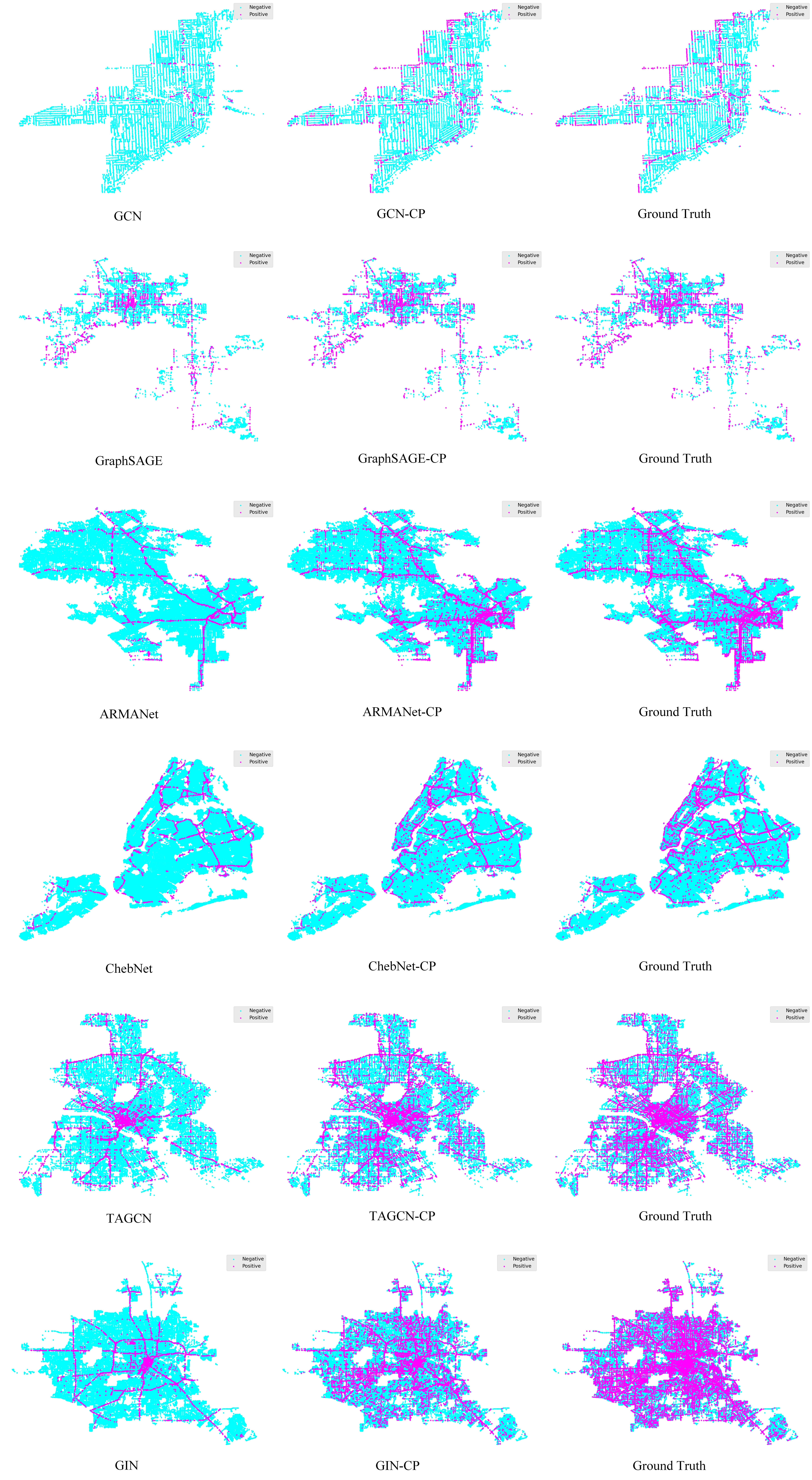}
    \vspace{-0.2cm}
    \caption{The visualization of the prediction by different methods on cities. Left columns: The original GNN methods. Middle columns: GNN methods with our proposed concurrency prior. Right columns: The ground truth.}

    \label{fig:Results_Part1}
\end{figure*}

\begin{figure*}[]
    \centering
    \includegraphics[width=0.7\textwidth]{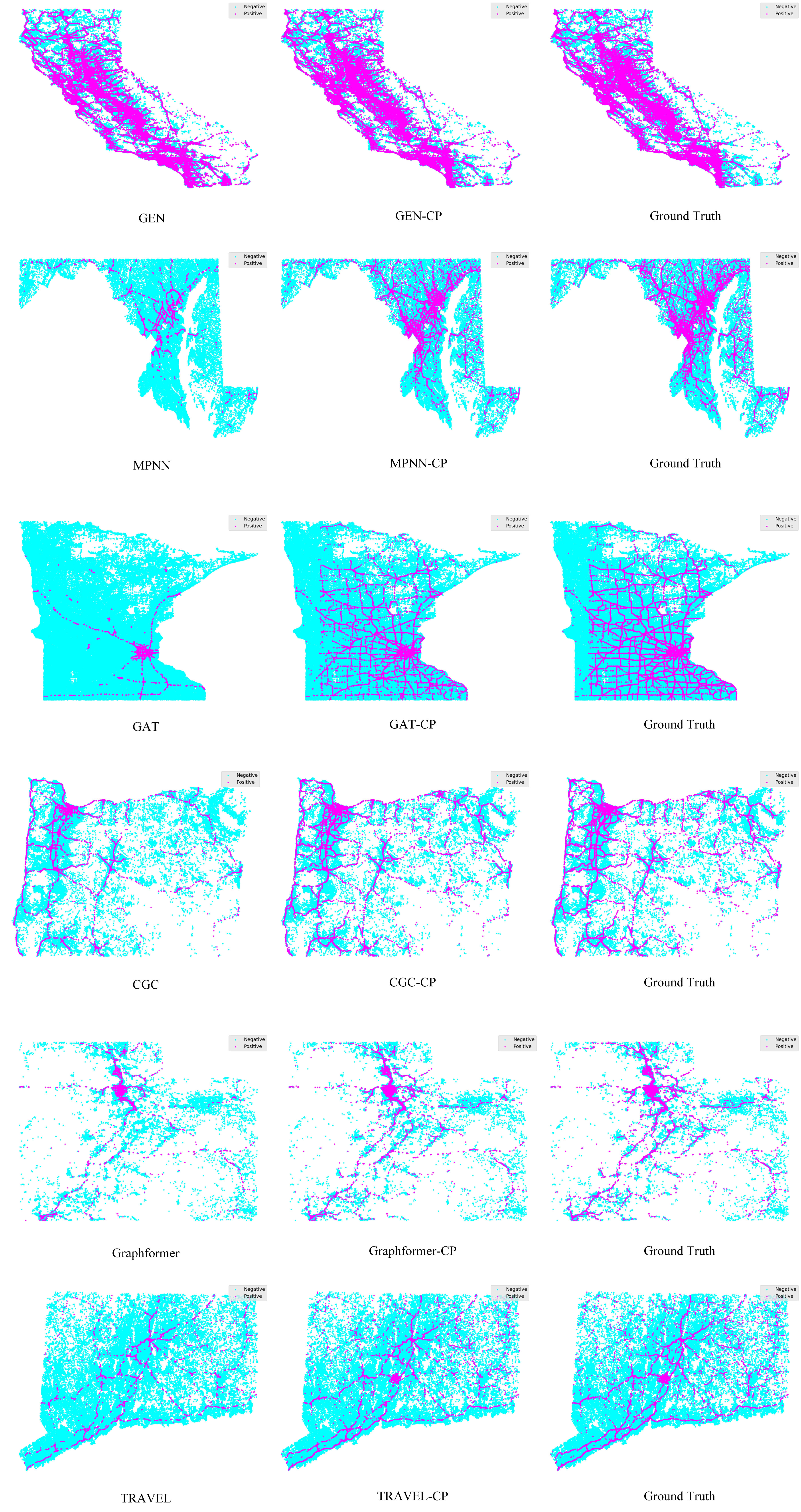}
    \vspace{-0.2cm}
    \caption{The visualization of the prediction by different methods on states. Left columns: The original GNN methods. Middle columns: GNN methods with our proposed concurrency prior. Right columns: The ground truth.}

    \label{fig:Results_Part2}
\end{figure*}

\noindent\textbf{Main Results.}
The numerical results are shown in Table \ref{tab:city-cls} and Table \ref{tab:state-cls}. The main observation is that by imposing our proposed  Concurrency Prior, all twelve GNN methods exhibit a considerable improvement across the cities and states. Specifically, in the city-wise datasets shown in Table \ref{tab:city-cls}, GNN can obtain a 1.31\% to 12.05\% gain in F1 score and a 1.81\% to 5.48\% gain in AUC by imposing our prior. We note that GCN and its early variants, including ChebNet, ARMANet, GraphSAGE, TAGCN, GIN, and GAT are significantly boosted to 46.58\%, 44.94\%, 42.99\%, 48.06\%, 46.82\%, and 45.24\% in F1 score, resulting in their performance being comparable with the original version of Graphformer (45.13\%). More importantly, our methods can still improve the previous best method (TRAVEL) with 1.31\% and 2.01\% gain in F1 score and AUC, respectively. Likewise, as shown in Table \ref{tab:state-cls}, our method can consistently enhance all GNN methods in state-wise datasets, which is on a larger geographic scale. We can observe a 3.26\% to 13.67\% and 1.33\% to 9.07\% gain in F1 and AUC, respectively. Similarly, the previous best method, TRAVEL, is boosted to 90.96\% in AUC, which is a high enough performance in such unbalanced datasets. These results underscore the value of integrating the proposed Concurrency Prior enhancements into GNNs for traffic incident prediction tasks. The visualization of the prediction for different GNN models on different cities and states is presented in Figs. \ref{fig:Results_Part1} and \ref{fig:Results_Part2}, showcasing a remarkable visual enhancement. Due to the page limit, we provide more visualization in our GitHub repository: \url{https://github.com/xiwenc1/Incident-GNN-CP}.

\begin{table*}[htbp]
\centering
\caption{City-wise accident occurrence prediction results in terms of F1 score and AUC. $\Delta$ denotes the gain obtained by imposing our proposed Concurrency Prior (with the suffix "-CP") in the neural networks.}
\label{tab:city-cls}
\resizebox{0.9\textwidth}{!}{%
\begin{tabular}{l|ll|ccccccccccccc}\toprule
\multicolumn{1}{c}{\textbf{Dataset}} & \multicolumn{2}{c}{\textbf{Mean}} & \multicolumn{2}{c}{\textbf{Miami   (FL)}} & \multicolumn{2}{c}{\textbf{Los   Angeles (CA)}} & \multicolumn{2}{c}{\textbf{Orlando   (FL)}} & \multicolumn{2}{c}{\textbf{Dallas   (TX)}} & \multicolumn{2}{c}{\textbf{Houston   (TX)}} & \multicolumn{2}{c}{\textbf{New   York (NY)}} \\
\multicolumn{1}{c}{\textbf{Method}} & \multicolumn{1}{c}{\textbf{F1}} & \multicolumn{1}{c}{\textbf{AUC}} & \multicolumn{1}{c}{\textbf{F1}} & \multicolumn{1}{c}{\textbf{AUC}} & \multicolumn{1}{c}{\textbf{F1}} & \multicolumn{1}{c}{\textbf{AUC}} & \multicolumn{1}{c}{\textbf{F1}} & \multicolumn{1}{c}{\textbf{AUC}} & \multicolumn{1}{c}{\textbf{F1}} & \multicolumn{1}{c}{\textbf{AUC}} & \multicolumn{1}{c}{\textbf{F1}} & \multicolumn{1}{c}{\textbf{AUC}} & \multicolumn{1}{c}{\textbf{F1}} & \multicolumn{1}{c}{\textbf{AUC}} \\ \toprule
GCN & 34.53 & 72.88 & 20.0$\pm$3.3 & 68.5$\pm$3.3 & 40.2$\pm$1.1 & 80.4$\pm$0.3 & 51.6$\pm$0.8 & 73.1$\pm$1.2 & 39.8$\pm$1.9 & 73.1$\pm$0.4 & 16.4$\pm$1.3 & 66.7$\pm$0.2 & 39.2$\pm$3.7 & 75.5$\pm$0.4 \\
GCN-CP & 46.58 & 76.32 & 46.49$\pm$2.83 & 80.35$\pm$0.88 & 53.68$\pm$1.57 & 83.56$\pm$0.17 & 56.38$\pm$3.15 & 77.40$\pm$1.15 & 45.15$\pm$1.30 & 71.86$\pm$0.41 & 31.63$\pm$1.21 & 63.90$\pm$0.22 & 46.13$\pm$0.81 & 80.83$\pm$0.25 \\
\rowcolor{lightgray}$\Delta$ & \textbf{+12.05} & \textbf{+3.44} & +26.49 & +11.85 & +13.48 & +3.16 & +4.78 & +4.30 & +5.35 & -1.24 & +15.23 & -2.80 & +6.93 & +5.33 \\
ChebNet & 36.72 & 75.45 & 20.7$\pm$2.9 & 71.3$\pm$3.6 & 39.8$\pm$1.8 & 81.0$\pm$0.3 & 53.1$\pm$0.6 & 76.7$\pm$1.6 & 42.0$\pm$0.5 & 75.8$\pm$0.4 & 23.8$\pm$0.5 & 69.6$\pm$0.5 & 40.9$\pm$4.3 & 78.3$\pm$1.1 \\
ChebNet-CP & 44.94 & 79.61 & 41.39$\pm$3.05 & 81.40$\pm$1.10 & 53.32$\pm$1.38 & 85.59$\pm$0.37 & 60.71$\pm$1.74 & 80.44$\pm$0.65 & 47.02$\pm$0.73 & 77.05$\pm$0.38 & 22.41$\pm$0.57 & 70.15$\pm$0.07 & 44.81$\pm$1.38 & 83.01$\pm$0.32 \\
\rowcolor{lightgray}$\Delta$ & \textbf{+8.22} & \textbf{+4.16} & +20.69 & +10.10 & +13.52 & +4.59 & +7.61 & +3.74 & +5.02 & +1.25 & -1.39 & +0.55 & +3.91 & +4.71 \\
ARMANet & 36.37 & 74.77 & 19.2$\pm$3.3 & 69.5$\pm$3.5 & 40.8$\pm$1.0 & 80.9$\pm$0.4 & 51.5$\pm$1.3 & 75.7$\pm$1.4 & 41.2$\pm$0.5 & 75.6$\pm$0.2 & 23.1$\pm$0.4 & 69.2$\pm$0.7 & 42.4$\pm$1.1 & 77.7$\pm$0.6 \\
ARMANet-CP & 44.88 & 79.54 & 42.30$\pm$5.06 & 81.58$\pm$1.34 & 52.78$\pm$1.84 & 85.41$\pm$0.13 & 59.82$\pm$1.63 & 80.15$\pm$1.00 & 47.31$\pm$1.49 & 77.03$\pm$0.48 & 23.56$\pm$1.79 & 70.16$\pm$0.16 & 43.50$\pm$1.49 & 82.89$\pm$0.42 \\
\rowcolor{lightgray}$\Delta$ & \textbf{+8.51} & \textbf{+4.77} & +23.10 & +12.08 & +11.98 & +4.51 & +8.32 & +4.45 & +6.11 & +1.43 & +0.46 & +0.96 & +1.10 & +5.19 \\
GraphSAGE & 37.55 & 73.57 & 20.7$\pm$2.4 & 67.6$\pm$2.8 & 41.6$\pm$0.5 & 80.5$\pm$0.3 & 52.6$\pm$1.3 & 74.1$\pm$1.2 & 44.2$\pm$0.5 & 74.4$\pm$0.3 & 23.7$\pm$0.4 & 68.5$\pm$0.4 & 42.5$\pm$1.1 & 76.3$\pm$0.1 \\
GraphSAGE-CP & 42.99 & 79.05 & 38.27$\pm$4.18 & 80.26$\pm$0.56 & 53.16$\pm$1.51 & 84.99$\pm$0.20 & 58.50$\pm$2.03 & 80.14$\pm$0.52 & 45.16$\pm$1.88 & 76.56$\pm$0.54 & 18.96$\pm$2.79 & 69.63$\pm$0.20 & 43.89$\pm$1.68 & 82.72$\pm$0.40 \\
\rowcolor{lightgray}$\Delta$ & \textbf{+5.44} & \textbf{+5.48} & +17.57 & +12.66 & +11.56 & +4.49 & +5.90 & +6.04 & +0.96 & +2.16 & -4.74 & +1.13 & +1.39 & +6.42 \\
TAGCN & 39.85 & 77.40 & 25.2$\pm$1.1 & 73.5$\pm$2.4 & 49.5$\pm$0.7 & 84.7$\pm$0.2 & 53.3$\pm$2.5 & 77.2$\pm$1.2 & 45.4$\pm$0.4 & 77.0$\pm$0.5 & 23.7$\pm$0.6 & 70.5$\pm$0.3 & 42.0$\pm$1.1 & 81.5$\pm$0.2 \\
TAGCN-CP & 48.06 & 82.38 & 45.83$\pm$2.24 & 85.44$\pm$0.59 & 56.61$\pm$0.97 & 88.67$\pm$0.30 & 62.42$\pm$1.68 & 82.67$\pm$0.58 & 49.11$\pm$0.99 & 79.00$\pm$0.35 & 26.81$\pm$2.10 & 71.91$\pm$0.18 & 47.59$\pm$1.20 & 86.58$\pm$0.24 \\
\rowcolor{lightgray}$\Delta$ & \textbf{+8.21} & \textbf{+4.98} & +20.63 & +11.94 & +7.11 & +3.97 & +9.12 & +5.47 & +3.71 & +2.00 & +3.11 & +1.41 & +5.59 & +5.08 \\
GIN & 37.17 & 75.57 & 22.8$\pm$1.2 & 72.7$\pm$2.6 & 41.6$\pm$0.7 & 81.8$\pm$0.2 & 54.7$\pm$1.4 & 76.6$\pm$1.1 & 41.3$\pm$2.0 & 75.2$\pm$0.3 & 20.9$\pm$1.0 & 68.0$\pm$0.3 & 41.7$\pm$2.1 & 79.1$\pm$0.5 \\
GIN-CP & 46.82 & 77.38 & 43.48$\pm$3.18 & 80.75$\pm$1.45 & 54.87$\pm$1.43 & 85.03$\pm$0.18 & 57.68$\pm$1.59 & 78.03$\pm$0.54 & 47.44$\pm$2.02 & 73.00$\pm$0.55 & 30.91$\pm$1.29 & 64.24$\pm$0.24 & 46.55$\pm$0.91 & 83.25$\pm$0.16 \\
\rowcolor{lightgray}$\Delta$ & \textbf{+9.65} & \textbf{+1.81} & +20.68 & +8.05 & +13.27 & +3.23 & +2.98 & +1.43 & +6.14 & -2.20 & +10.01 & -3.76 & +4.85 & +4.15 \\
GAT & 36.93 & 73.47 & 22.6$\pm$1.5 & 68.3$\pm$3.0 & 41.6$\pm$0.4 & 80.9$\pm$0.2 & 55.3$\pm$1.3 & 74.1$\pm$1.0 & 42.1$\pm$1.5 & 73.6$\pm$0.3 & 17.8$\pm$0.8 & 67.3$\pm$0.3 & 42.2$\pm$0.5 & 76.6$\pm$0.4 \\
GAT-CP & 45.24 & 76.06 & 40.07$\pm$10.92 & 78.96$\pm$1.70 & 50.71$\pm$1.09 & 83.11$\pm$0.19 & 55.09$\pm$6.56 & 76.83$\pm$0.65 & 47.07$\pm$1.38 & 71.76$\pm$0.14 & 35.14$\pm$0.90 & 65.29$\pm$0.33 & 43.36$\pm$3.32 & 80.41$\pm$0.63 \\
\rowcolor{lightgray}$\Delta$ & \textbf{+8.31} & \textbf{+2.59 }& +17.47 & +10.66 & +9.11 & +2.21 & -0.21 & +2.73 & +4.97 & -1.84 & +17.34 & -2.01 & +1.16 & +3.81 \\
MPNN & 44.63 & 81.32 & 38.8$\pm$2.1 & 82.4$\pm$1.0 & 46.0$\pm$1.6 & 83.9$\pm$0.2 & 61.4$\pm$2.5 & 81.8$\pm$0.7 & 48.5$\pm$1.9 & 79.4$\pm$0.4 & 28.2$\pm$1.7 & 73.5$\pm$0.5 & 44.9$\pm$0.8 & 86.9$\pm$0.4 \\
MPNN-CP & 45.36 & 83.50 & 39.56$\pm$5.44 & 86.22$\pm$0.32 & 51.72$\pm$4.24 & 87.79$\pm$0.08 & 63.09$\pm$1.39 & 84.03$\pm$0.28 & 47.93$\pm$1.54 & 80.33$\pm$0.38 & 23.34$\pm$2.26 & 73.82$\pm$0.27 & 46.56$\pm$3.49 & 88.83$\pm$0.57 \\
\rowcolor{lightgray}$\Delta$ & \textbf{+0.73} & \textbf{+2.18} & +0.76 & +3.82 & +5.72 & +3.89 & +1.69 & +2.23 & -0.57 & +0.93 & -4.86 & +0.32 & +1.66 & +1.93 \\
CGC & 42.47 & 79.83 & 34.4$\pm$2.7 & 79.5$\pm$1.5 & 45.0$\pm$1.2 & 81.5$\pm$0.2 & 59.0$\pm$2.1 & 81.1$\pm$0.8 & 48.5$\pm$0.5 & 79.2$\pm$0.7 & 27.3$\pm$1.9 & 72.3$\pm$0.1 & 40.6$\pm$1.2 & 85.4$\pm$0.8 \\
CGC-CP & 48.55 & 82.84 & 42.94$\pm$3.34 & 86.18$\pm$0.44 & 53.55$\pm$1.37 & 87.08$\pm$0.21 & 64.19$\pm$1.88 & 83.68$\pm$0.73 & 48.12$\pm$5.22 & 79.69$\pm$0.16 & 35.68$\pm$6.53 & 72.39$\pm$0.24 & 46.82$\pm$3.02 & 88.00$\pm$0.62 \\
\rowcolor{lightgray}$\Delta$ & \textbf{+6.08} & \textbf{+3.01} & +8.54 & +6.68 & +8.55 & +5.58 & +5.19 & +2.58 & -0.38 & +0.49 & +8.38 & +0.09 & +6.22 & +2.60 \\
Graphformer & 45.13 & 81.32 & 37.7$\pm$3.3 & 81.0$\pm$1.9 & 48.9$\pm$0.3 & 83.8$\pm$0.3 & 62.9$\pm$1.6 & 82.0$\pm$0.7 & 49.8$\pm$0.7 & 80.0$\pm$0.7 & 28.4$\pm$0.7 & 73.9$\pm$0.4 & 43.1$\pm$0.7 & 87.2$\pm$0.4 \\
Graphformer-CP & 50.14 & 83.36 & 49.18$\pm$2.74 & 85.58$\pm$0.80 & 54.82$\pm$1.53 & 86.93$\pm$0.26 & 66.08$\pm$0.66 & 84.19$\pm$0.56 & 51.76$\pm$0.93 & 80.59$\pm$0.36 & 31.01$\pm$1.91 & 74.04$\pm$0.24 & 47.97$\pm$1.75 & 88.85$\pm$0.47 \\
\rowcolor{lightgray}$\Delta$ & \textbf{+5.01} & \textbf{+2.04} & +11.48 & +4.58 & +5.92 & +3.13 & +3.18 & +2.19 & +1.96 & +0.59 & +2.61 & +0.14 & +4.87 & +1.65 \\
GEN & 49.07 & 80.97 & 44.9$\pm$3.1 & 81.0$\pm$2.4 & 48.6$\pm$6.2 & 82.7$\pm$0.9 & 63.0$\pm$1.1 & 81.2$\pm$0.9 & 56.5$\pm$1.7 & 79.5$\pm$0.1 & 34.1$\pm$6.0 & 73.7$\pm$0.4 & 47.3$\pm$1.4 & 87.7$\pm$0.9 \\
GEN-CP & 53.12 & 83.55 & 49.43$\pm$1.39 & 85.56$\pm$0.61 & 57.88$\pm$1.07 & 88.05$\pm$0.62 & 66.18$\pm$1.81 & 83.89$\pm$0.34 & 51.47$\pm$5.09 & 80.29$\pm$0.08 & 40.62$\pm$6.89 & 74.28$\pm$0.21 & 53.12$\pm$0.78 & 89.21$\pm$0.23 \\
\rowcolor{lightgray}$\Delta$ & \textbf{+4.05} & \textbf{+2.58} & +4.53 & +4.56 & +9.28 & +5.35 & +3.18 & +2.69 & -5.03 & +0.79 & +6.52 & +0.58 & +5.82 & +1.51 \\
TRAVEL & 54.62 & 82.77 & 51.9$\pm$1.0 & 84.9$\pm$0.9 & 55.3$\pm$0.9 & 85.9$\pm$0.5 & 65.0$\pm$0.4 & 82.3$\pm$0.4 & 58.0$\pm$0.9 & 80.8$\pm$0.7 & 46.4$\pm$0.7 & 74.5$\pm$0.3 & 51.1$\pm$0.9 & 88.2$\pm$0.2 \\
TRAVEL-CP & 55.93 & 84.78 & 56.84$\pm$2.57 & 88.05$\pm$0.41 & 59.56$\pm$1.11 & 89.02$\pm$0.44 & 67.85$\pm$0.97 & 85.15$\pm$0.58 & 56.63$\pm$1.23 & 81.35$\pm$0.24 & 43.55$\pm$4.49 & 75.59$\pm$0.22 & 51.18$\pm$1.56 & 89.52$\pm$0.07 \\
\rowcolor{lightgray}$\Delta$ & \textbf{+1.31} & \textbf{+2.01} & +4.94 & +3.15 & +4.26 & +3.12 & +2.85 & +2.85 & -1.37 & +0.55 & -2.85 & +1.09 & +0.08 & +1.32 \\ \bottomrule
\end{tabular}%
}
\end{table*}

\begin{table*}[]
\centering
\caption{State-wise accident occurrence prediction results in terms of F1 score and AUC. $\Delta$ denotes the gain obtained by imposing our proposed Concurrency Prior (with the suffix "-CP") in the neural networks.}
\label{tab:state-cls}
\resizebox{0.9\textwidth}{!}{%
\begin{tabular}{l|ll|ccccccccccccc}\toprule
\multicolumn{1}{c}{\textbf{Dataset}} & \multicolumn{2}{c}{\textbf{Mean}} & \multicolumn{2}{c}{\textbf{California}} & \multicolumn{2}{c}{\textbf{Oregon}} & \multicolumn{2}{c}{\textbf{Utah}} & \multicolumn{2}{c}{\textbf{Maryland}} & \multicolumn{2}{c}{\textbf{Minnesota}} & \multicolumn{2}{c}{\textbf{Connecticut}} \\ 
\multicolumn{1}{c}{\textbf{Method}} & \multicolumn{1}{c}{\textbf{F1}} & \multicolumn{1}{c}{\textbf{AUC}} & \multicolumn{1}{c}{\textbf{F1}} & \multicolumn{1}{c}{\textbf{AUC}} & \multicolumn{1}{c}{\textbf{F1}} & \multicolumn{1}{c}{\textbf{AUC}} & \multicolumn{1}{c}{\textbf{F1}} & \multicolumn{1}{c}{\textbf{AUC}} & \multicolumn{1}{c}{\textbf{F1}} & \multicolumn{1}{c}{\textbf{AUC}} & \multicolumn{1}{c}{\textbf{F1}} & \multicolumn{1}{c}{\textbf{AUC}} & \multicolumn{1}{c}{\textbf{F1}} & \multicolumn{1}{c}{\textbf{AUC}} \\ \toprule
GCN & 28.58 & 73.85 & 24.0$\pm$0.0 & 71.5$\pm$0.0 & 20.6$\pm$0.5 & 68.7$\pm$0.7 & 32.7$\pm$0.1 & 76.3$\pm$0.3 & 26.1$\pm$1.3 & 79.5$\pm$0.4 & 28.1$\pm$0.4 & 70.9$\pm$0.3 & 40.0$\pm$0.6 & 76.2$\pm$0.8 \\
GCN-CP & 42.25 & 81.05 & 45.61$\pm$1.20 & 79.24$\pm$0.21 & 45.58$\pm$0.83 & 81.48$\pm$0.15 & 45.54$\pm$0.90 & 83.56$\pm$0.25 & 36.00$\pm$1.27 & 80.90$\pm$0.27 & 41.90$\pm$1.47 & 81.06$\pm$0.19 & 38.86$\pm$3.60 & 80.05$\pm$0.34 \\
\rowcolor{lightgray}$\Delta$ & \textbf{+13.67} & \textbf{+7.20} & +21.61 & +7.74 & +24.98 & +12.78 & +12.84 & +7.26 & +9.90 & +1.40 & +13.80 & +10.16 & -1.14 & +3.85 \\
ChebNet & 29.87 & 75.73 & 23.2$\pm$1.0 & 72.9$\pm$0.2 & 21.0$\pm$0.2 & 73.1$\pm$0.3 & 34.3$\pm$1.2 & 77.3$\pm$0.5 & 28.5$\pm$0.3 & 80.4$\pm$0.1 & 30.2$\pm$2.1 & 74.1$\pm$1.3 & 42.0$\pm$0.4 & 76.6$\pm$0.2 \\
ChebNet-CP & 38.74 & 82.58 & 38.66$\pm$0.69 & 81.09$\pm$0.11 & 35.68$\pm$9.84 & 82.72$\pm$0.10 & 42.78$\pm$1.81 & 84.46$\pm$0.18 & 35.20$\pm$1.35 & 83.61$\pm$0.16 & 38.88$\pm$1.35 & 82.11$\pm$0.11 & 41.25$\pm$1.37 & 81.49$\pm$0.19 \\
\rowcolor{lightgray}$\Delta$ & \textbf{+8.87} & \textbf{+6.85} & +15.46 & +8.19 & +14.68 & +9.62 & +8.48 & +7.16 & +6.70 & +3.21 & +8.68 & +8.01 & -0.75 & +4.89 \\
ARMANet & 29.03 & 75.53 & 23.6$\pm$2.0 & 72.8$\pm$0.2 & 18.6$\pm$3.4 & 72.7$\pm$0.7 & 34.6$\pm$0.3 & 77.2$\pm$0.3 & 28.6$\pm$1.6 & 80.6$\pm$0.2 & 26.4$\pm$1.7 & 72.7$\pm$1.2 & 42.4$\pm$1.5 & 77.2$\pm$0.6 \\
ARMANet-CP & 38.45 & 82.60 & 38.48$\pm$0.58 & 81.07$\pm$0.05 & 38.49$\pm$1.33 & 82.71$\pm$0.21 & 44.04$\pm$1.83 & 84.82$\pm$0.35 & 33.54$\pm$1.66 & 83.48$\pm$0.12 & 37.57$\pm$1.55 & 81.96$\pm$0.08 & 38.56$\pm$1.19 & 81.54$\pm$0.24 \\
\rowcolor{lightgray}$\Delta$ & \textbf{+9.42} &\textbf{+7.07} & +14.88 & +8.27 & +19.89 & +10.01 & +9.44 & +7.62 & +4.94 & +2.88 & +11.17 & +9.26 & -3.84 & +4.34 \\
GraphSAGE & 30.27 & 75.17 & 25.8$\pm$0.4 & 72.8$\pm$0.4 & 21.4$\pm$0.9 & 71.2$\pm$1.3 & 34.3$\pm$1.5 & 77.7$\pm$0.5 & 28.5$\pm$1.2 & 80.2$\pm$0.1 & 28.9$\pm$0.1 & 71.9$\pm$0.8 & 42.7$\pm$1.6 & 77.2$\pm$0.6 \\
GraphSAGE-CP & 37.67 & 82.23 & 39.52$\pm$0.51 & 81.01$\pm$0.09 & 41.53$\pm$4.20 & 82.32$\pm$0.12 & 41.23$\pm$5.77 & 83.93$\pm$0.18 & 30.64$\pm$8.28 & 82.83$\pm$0.06 & 34.91$\pm$3.06 & 81.50$\pm$0.23 & 38.18$\pm$1.73 & 81.78$\pm$0.39 \\
\rowcolor{lightgray}$\Delta$ & \textbf{+7.40} & \textbf{+7.06} & +13.72 & +8.21 & +20.13 & +11.12 & +6.93 & +6.23 & +2.14 & +2.63 & +6.01 & +9.60 & -4.52 & +4.58 \\
TAGCN & 30.37 & 78.10 & 28.7$\pm$0.4 & 75.9$\pm$0.1 & 24.7$\pm$1.0 & 76.6$\pm$0.1 & 34.1$\pm$1.0 & 78.9$\pm$0.4 & 26.1$\pm$0.9 & 81.8$\pm$0.4 & 30.8$\pm$0.9 & 77.2$\pm$0.8 & 37.8$\pm$0.8 & 78.2$\pm$1.3 \\
TAGCN-CP & 43.13 & 87.17 & 47.25$\pm$1.50 & 85.81$\pm$0.15 & 46.08$\pm$1.80 & 88.67$\pm$0.12 & 46.79$\pm$1.34 & 88.97$\pm$0.30 & 36.78$\pm$1.29 & 86.67$\pm$0.28 & 42.40$\pm$2.09 & 87.47$\pm$0.07 & 39.49$\pm$1.53 & 85.43$\pm$0.47 \\
\rowcolor{lightgray}$\Delta$ & \textbf{+12.76} & \textbf{+9.07} & +18.55 & +9.91 & +21.38 & +12.07 & +12.69 & +10.07 & +10.68 & +4.87 & +11.60 & +10.27 & +1.69 & +7.23 \\
GIN & 31.78 & 76.45 & 28.0$\pm$0.2 & 72.7$\pm$0.2 & 24.3$\pm$0.4 & 74.2$\pm$0.2 & 36.2$\pm$0.3 & 78.9$\pm$0.5 & 28.2$\pm$0.6 & 80.8$\pm$0.2 & 32.0$\pm$1.8 & 74.9$\pm$1.3 & 42.0$\pm$1.2 & 77.2$\pm$0.4 \\
GIN-CP & 43.97 & 82.08 & 45.42$\pm$1.83 & 79.76$\pm$0.05 & 48.36$\pm$2.03 & 83.07$\pm$0.34 & 46.60$\pm$1.21 & 84.41$\pm$0.19 & 37.38$\pm$1.64 & 82.23$\pm$0.18 & 44.34$\pm$1.76 & 82.35$\pm$0.12 & 41.73$\pm$1.87 & 80.67$\pm$0.36 \\
\rowcolor{lightgray}$\Delta$ & \textbf{+12.19} & \textbf{+5.63} & +17.42 & +7.06 & +24.06 & +8.87 & +10.40 & +5.51 & +9.18 & +1.43 & +12.34 & +7.45 & -0.27 & +3.47 \\
GAT & 30.12 & 74.47 & 24.5$\pm$0.3 & 72.2$\pm$0.1 & 21.8$\pm$0.4 & 70.2$\pm$1.0 & 35.6$\pm$2.2 & 76.8$\pm$0.6 & 27.8$\pm$0.2 & 80.4$\pm$0.2 & 29.1$\pm$0.2 & 71.3$\pm$1.1 & 41.9$\pm$2.6 & 75.9$\pm$1.5 \\
GAT-CP & 40.06 & 80.83 & 46.70$\pm$0.59 & 79.76$\pm$0.21 & 47.12$\pm$0.79 & 81.26$\pm$0.57 & 39.62$\pm$6.36 & 82.76$\pm$1.13 & 36.07$\pm$0.85 & 81.61$\pm$0.75 & 38.14$\pm$9.85 & 80.81$\pm$0.52 & 32.69$\pm$2.74 & 78.77$\pm$0.22 \\
\rowcolor{lightgray}$\Delta$ & \textbf{+9.94} & \textbf{+6.36} & +22.20 & +7.56 & +25.32 & +11.06 & +4.02 & +5.96 & +8.27 & +1.21 & +9.04 & +9.51 & -9.21 & +2.87 \\
MPNN & 41.20 & 87.13 & 33.2$\pm$1.7 & 79.9$\pm$0.5 & 56.2$\pm$1.3 & 89.2$\pm$0.8 & 43.1$\pm$2.6 & 88.7$\pm$0.3 & 32.3$\pm$0.2 & 89.4$\pm$0.2 & 38.5$\pm$1.0 & 89.8$\pm$0.1 & 43.9$\pm$0.6 & 85.8$\pm$0.7 \\
MPNN-CP & 44.93 & 89.50 & 43.84$\pm$1.93 & 84.45$\pm$0.23 & 57.06$\pm$1.95 & 90.83$\pm$0.17 & 45.30$\pm$4.70 & 91.59$\pm$0.12 & 34.03$\pm$2.50 & 90.38$\pm$0.08 & 45.99$\pm$2.09 & 91.69$\pm$0.07 & 43.35$\pm$2.63 & 88.07$\pm$0.26 \\
\rowcolor{lightgray}$\Delta$ & \textbf{+3.73} & \textbf{+2.37} & +10.64 & +4.55 & +0.86 & +1.63 & +2.20 & +2.89 & +1.73 & +0.98 & +7.49 & +1.89 & -0.55 & +2.27 \\
CGC & 44.80 & 86.50 & 34.4$\pm$2.8 & 79.0$\pm$0.1 & 53.1$\pm$3.3 & 88.5$\pm$0.1 & 47.1$\pm$0.8 & 88.3$\pm$0.2 & 43.0$\pm$1.9 & 88.6$\pm$0.4 & 48.1$\pm$1.7 & 88.7$\pm$0.5 & 43.1$\pm$0.2 & 85.9$\pm$0.7 \\
CGC-CP & 48.18 & 88.80 & 46.02$\pm$2.08 & 84.10$\pm$0.06 & 58.00$\pm$2.53 & 90.34$\pm$0.16 & 52.05$\pm$3.34 & 90.89$\pm$0.11 & 41.65$\pm$4.13 & 89.84$\pm$0.07 & 48.01$\pm$5.12 & 90.74$\pm$0.17 & 43.34$\pm$3.75 & 86.89$\pm$0.40 \\
\rowcolor{lightgray}$\Delta$ & \textbf{+3.38} & \textbf{+2.30} & +11.62 & +5.10 & +4.90 & +1.84 & +4.95 & +2.59 & -1.35 & +1.24 & -0.09 & +2.04 & +0.24 & +0.99 \\
Graphformer & 39.48 & 87.15 & 29.3$\pm$0.5 & 80.1$\pm$0.1 & 53.5$\pm$0.4 & 89.4$\pm$0.3 & 44.8$\pm$1.4 & 89.0$\pm$0.1 & 31.5$\pm$0.6 & 89.2$\pm$0.6 & 33.9$\pm$0.5 & 90.1$\pm$0.2 & 43.9$\pm$0.7 & 85.1$\pm$0.6 \\
Graphformer-CP & 47.77 & 88.48 & 44.68$\pm$0.38 & 84.25$\pm$0.11 & 56.96$\pm$1.04 & 89.71$\pm$0.14 & 53.18$\pm$1.72 & 90.35$\pm$0.25 & 38.97$\pm$2.81 & 89.33$\pm$0.14 & 49.43$\pm$1.83 & 90.13$\pm$0.09 & 43.43$\pm$1.37 & 87.13$\pm$0.21 \\
\rowcolor{lightgray}$\Delta$ & \textbf{+8.29} & \textbf{+1.33} & +15.38 & +4.15 & +3.46 & +0.31 & +8.38 & +1.35 & +7.47 & +0.13 & +15.53 & +0.03 & -0.47 & +2.03 \\
GEN & 48.88 & 84.92 & 43.9$\pm$0.1 & 77.3$\pm$0.4 & 59.4$\pm$0.8 & 87.2$\pm$0.7 & 53.8$\pm$0.8 & 86.1$\pm$0.3 & 40.8$\pm$2.8 & 88.4$\pm$0.8 & 52.3$\pm$1.5 & 88.7$\pm$0.4 & 43.1$\pm$1.1 & 81.8$\pm$0.7 \\
GEN-CP & 54.30 & 89.81 & 53.30$\pm$1.70 & 86.03$\pm$0.10 & 62.40$\pm$0.80 & 91.28$\pm$0.09 & 56.93$\pm$1.14 & 91.28$\pm$0.24 & 46.44$\pm$3.35 & 90.03$\pm$0.19 & 56.77$\pm$2.28 & 91.71$\pm$0.37 & 49.94$\pm$1.11 & 88.53$\pm$0.49 \\
\rowcolor{lightgray}$\Delta$ & \textbf{+5.42} & \textbf{+4.89} & +9.40 & +8.73 & +3.00 & +4.08 & +3.13 & +5.18 & +5.64 & +1.63 & +4.47 & +3.01 & +6.84 & +6.73 \\
TRAVEL & 51.78 & 88.02 & 46.1$\pm$0.7 & 81.1$\pm$1.0 & 60.8$\pm$0.0 & 90.5$\pm$0.2 & 55.6$\pm$0.6 & 89.9$\pm$1.9 & 46.4$\pm$1.4 & 90.1$\pm$0.3 & 55.2$\pm$1.0 & 91.3$\pm$0.7 & 46.6$\pm$0.3 & 85.2$\pm$0.6 \\
TRAVEL-CP & 55.04 & 90.96 & 54.85$\pm$0.58 & 87.22$\pm$0.18 & 62.46$\pm$1.03 & 92.24$\pm$0.23 & 57.42$\pm$1.84 & 92.49$\pm$0.22 & 47.44$\pm$1.17 & 90.99$\pm$0.16 & 58.93$\pm$0.25 & 92.87$\pm$0.17 & 49.17$\pm$1.62 & 89.93$\pm$0.38 \\
\rowcolor{lightgray}$\Delta$ & \textbf{+3.26} & \textbf{+2.94} & +8.75 & +6.12 & +1.66 & +1.74 & +1.82 & +2.59 & +1.04 & +0.89 & +3.73 & +1.57 & +2.57 & +4.73 \\ \bottomrule
\end{tabular}%
}
\end{table*}

\begin{figure}[]
    \centering
    \includegraphics[width=0.44\textwidth]{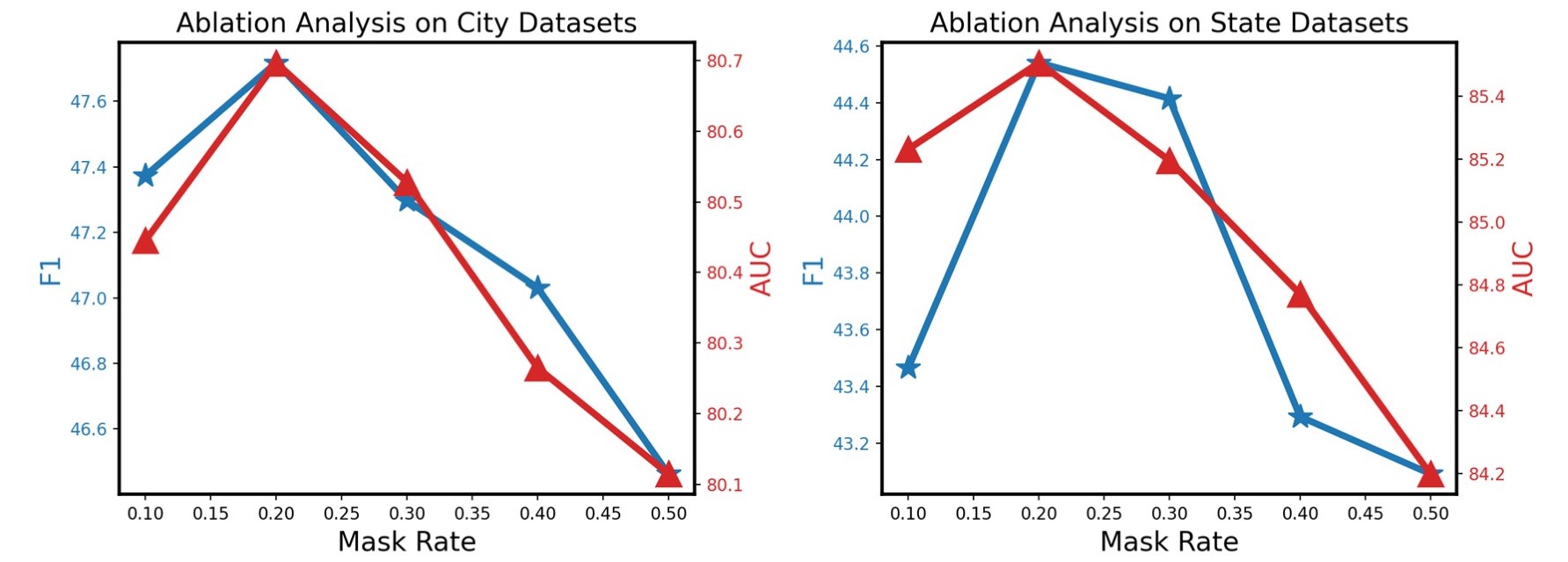}
    \vspace{-0.2cm}
    \caption{The ablation analysis for different masking rate.}

    \label{fig:aba}
\end{figure}

\noindent\textbf{Ablation Analysis.}
The only hyperparameter in our proposed method is the rate of masking; hence here, we conduct the experiment to test the effect by different $R\in (0,0.5)$. We present the average F1 and AUC across all selected states and cities in Fig. \ref{fig:aba}.

\section{Conclusion}
In this paper, we perform the traffic road network analysis using a large-scale graph-based nationwide data source, including incident records across 49 states in the USA.
we first propose two metrics, Average Neighbor Crash Density (ANCD) and Average Neighbor Crash Continuity (ANCC), to statically validate the intuitive concurrency hypothesis, where there is a high probability of incidents occurring in neighboring nodes of a road network. Based on this validation, we then propose our novel Concurrency Prior (CP) method that can incorporate this concurrency information into various GNN models with neglectable extra parameters. Our experiment showcases a remarkable improvement in the semi-supervised graph-based traffic incident prediction tasks.
We expect our contributions will be able to offer promising directions for future research and practical applications in urban planning and public safety.

\section*{Acknowledgements}
This material is based upon the work supported by the National Science Foundation under Grant Number 2204721 and partially supported by our collaborative project with MIT Lincoln Lab under Grant Number 7000612889.

\newpage
\bibliographystyle{ACM-Reference-Format}
\bibliography{main.bib}


\begin{thebibliography}{62}


\ifx \showCODEN    \undefined \def \showCODEN     #1{\unskip}     \fi
\ifx \showDOI      \undefined \def \showDOI       #1{#1}\fi
\ifx \showISBNx    \undefined \def \showISBNx     #1{\unskip}     \fi
\ifx \showISBNxiii \undefined \def \showISBNxiii  #1{\unskip}     \fi
\ifx \showISSN     \undefined \def \showISSN      #1{\unskip}     \fi
\ifx \showLCCN     \undefined \def \showLCCN      #1{\unskip}     \fi
\ifx \shownote     \undefined \def \shownote      #1{#1}          \fi
\ifx \showarticletitle \undefined \def \showarticletitle #1{#1}   \fi
\ifx \showURL      \undefined \def \showURL       {\relax}        \fi
\providecommand\bibfield[2]{#2}
\providecommand\bibinfo[2]{#2}
\providecommand\natexlab[1]{#1}
\providecommand\showeprint[2][]{arXiv:#2}

\bibitem[Barab{\'a}si and Albert(1999)]%
        {barabasi1999emergence}
\bibfield{author}{\bibinfo{person}{Albert-L{\'a}szl{\'o} Barab{\'a}si} {and} \bibinfo{person}{R{\'e}ka Albert}.} \bibinfo{year}{1999}\natexlab{}.
\newblock \showarticletitle{Emergence of scaling in random networks}.
\newblock \bibinfo{journal}{\emph{science}} \bibinfo{volume}{286}, \bibinfo{number}{5439} (\bibinfo{year}{1999}), \bibinfo{pages}{509--512}.
\newblock


\bibitem[Bianchi et~al\mbox{.}(2021)]%
        {bianchi2021armanet}
\bibfield{author}{\bibinfo{person}{Filippo~Maria Bianchi}, \bibinfo{person}{Daniele Grattarola}, \bibinfo{person}{Lorenzo Livi}, {and} \bibinfo{person}{Cesare Alippi}.} \bibinfo{year}{2021}\natexlab{}.
\newblock \showarticletitle{Graph Neural Networks with Convolutional ARMA Filters}.
\newblock \bibinfo{journal}{\emph{IEEE Transactions on Pattern Analysis and Machine Intelligence}} (\bibinfo{year}{2021}), \bibinfo{pages}{1–1}.
\newblock
\showISSN{1939-3539}
\urldef\tempurl%
\url{https://doi.org/10.1109/tpami.2021.3054830}
\showDOI{\tempurl}


\bibitem[Boccaletti et~al\mbox{.}(2014)]%
        {boccaletti2014structure}
\bibfield{author}{\bibinfo{person}{Stefano Boccaletti}, \bibinfo{person}{Ginestra Bianconi}, \bibinfo{person}{Regino Criado}, \bibinfo{person}{Charo~I Del~Genio}, \bibinfo{person}{Jes{\'u}s G{\'o}mez-Gardenes}, \bibinfo{person}{Miguel Romance}, \bibinfo{person}{Irene Sendina-Nadal}, \bibinfo{person}{Zhen Wang}, {and} \bibinfo{person}{Massimiliano Zanin}.} \bibinfo{year}{2014}\natexlab{}.
\newblock \showarticletitle{The structure and dynamics of multilayer networks}.
\newblock \bibinfo{journal}{\emph{Physics reports}} \bibinfo{volume}{544}, \bibinfo{number}{1} (\bibinfo{year}{2014}), \bibinfo{pages}{1--122}.
\newblock


\bibitem[Boeing(2017)]%
        {boeing2017osmnx}
\bibfield{author}{\bibinfo{person}{Geoff Boeing}.} \bibinfo{year}{2017}\natexlab{}.
\newblock \showarticletitle{OSMnx: New methods for acquiring, constructing, analyzing, and visualizing complex street networks}.
\newblock \bibinfo{journal}{\emph{Computers, Environment and Urban Systems}}  \bibinfo{volume}{65} (\bibinfo{year}{2017}), \bibinfo{pages}{126--139}.
\newblock


\bibitem[Bruna et~al\mbox{.}(2013)]%
        {bruna2013spectral}
\bibfield{author}{\bibinfo{person}{Joan Bruna}, \bibinfo{person}{Wojciech Zaremba}, \bibinfo{person}{Arthur Szlam}, {and} \bibinfo{person}{Yann LeCun}.} \bibinfo{year}{2013}\natexlab{}.
\newblock \showarticletitle{Spectral networks and locally connected networks on graphs}.
\newblock \bibinfo{journal}{\emph{arXiv preprint arXiv:1312.6203}} (\bibinfo{year}{2013}).
\newblock


\bibitem[Cai et~al\mbox{.}(2022)]%
        {cai2022dq}
\bibfield{author}{\bibinfo{person}{Peide Cai}, \bibinfo{person}{Hengli Wang}, \bibinfo{person}{Yuxiang Sun}, {and} \bibinfo{person}{Ming Liu}.} \bibinfo{year}{2022}\natexlab{}.
\newblock \showarticletitle{DQ-GAT: Towards safe and efficient autonomous driving with deep Q-learning and graph attention networks}.
\newblock \bibinfo{journal}{\emph{IEEE Transactions on Intelligent Transportation Systems}} \bibinfo{volume}{23}, \bibinfo{number}{11} (\bibinfo{year}{2022}), \bibinfo{pages}{21102--21112}.
\newblock


\bibitem[Caliendo et~al\mbox{.}(2007)]%
        {caliendo2007regressionaccident}
\bibfield{author}{\bibinfo{person}{Ciro Caliendo}, \bibinfo{person}{Maurizio Guida}, {and} \bibinfo{person}{Alessandra Parisi}.} \bibinfo{year}{2007}\natexlab{}.
\newblock \showarticletitle{A crash-prediction model for multilane roads}.
\newblock \bibinfo{journal}{\emph{Accident Analysis \& Prevention}} \bibinfo{volume}{39}, \bibinfo{number}{4} (\bibinfo{year}{2007}), \bibinfo{pages}{657--670}.
\newblock


\bibitem[Chen et~al\mbox{.}(2022)]%
        {chen2022network}
\bibfield{author}{\bibinfo{person}{Xiwen Chen}, \bibinfo{person}{Hao Wang}, \bibinfo{person}{Abolfazl Razi}, \bibinfo{person}{Brendan Russo}, \bibinfo{person}{Jason Pacheco}, \bibinfo{person}{John Roberts}, \bibinfo{person}{Jeffrey Wishart}, \bibinfo{person}{Larry Head}, {and} \bibinfo{person}{Alonso~Granados Baca}.} \bibinfo{year}{2022}\natexlab{}.
\newblock \showarticletitle{Network-level safety metrics for overall traffic safety assessment: A case study}.
\newblock \bibinfo{journal}{\emph{IEEE Access}}  \bibinfo{volume}{11} (\bibinfo{year}{2022}), \bibinfo{pages}{17755--17778}.
\newblock


\bibitem[Defferrard et~al\mbox{.}(2016)]%
        {defferrard2016chebnet}
\bibfield{author}{\bibinfo{person}{Micha{\"e}l Defferrard}, \bibinfo{person}{Xavier Bresson}, {and} \bibinfo{person}{Pierre Vandergheynst}.} \bibinfo{year}{2016}\natexlab{}.
\newblock \showarticletitle{Convolutional neural networks on graphs with fast localized spectral filtering}.
\newblock \bibinfo{journal}{\emph{Advances in neural information processing systems}}  \bibinfo{volume}{29} (\bibinfo{year}{2016}), \bibinfo{pages}{3844--3852}.
\newblock


\bibitem[Du et~al\mbox{.}(2018)]%
        {du2018tagcn}
\bibfield{author}{\bibinfo{person}{Jian Du}, \bibinfo{person}{Shanghang Zhang}, \bibinfo{person}{Guanhang Wu}, \bibinfo{person}{Jose M.~F. Moura}, {and} \bibinfo{person}{Soummya Kar}.} \bibinfo{year}{2018}\natexlab{}.
\newblock \bibinfo{title}{Topology Adaptive Graph Convolutional Networks}.
\newblock
\newblock
\showeprint[arxiv]{1710.10370}~[cs.LG]


\bibitem[Fadhel et~al\mbox{.}(2024)]%
        {fadhel2024comprehensive}
\bibfield{author}{\bibinfo{person}{Mohammed~A Fadhel}, \bibinfo{person}{Ali~M Duhaim}, \bibinfo{person}{Ahmed Saihood}, \bibinfo{person}{Ahmed Sewify}, \bibinfo{person}{Mokhaled~NA Al-Hamadani}, \bibinfo{person}{AS Albahri}, \bibinfo{person}{Laith Alzubaidi}, \bibinfo{person}{Ashish Gupta}, \bibinfo{person}{Sayedali Mirjalili}, {and} \bibinfo{person}{Yuantong Gu}.} \bibinfo{year}{2024}\natexlab{}.
\newblock \showarticletitle{Comprehensive Systematic Review of Information Fusion Methods in Smart Cities and Urban Environments}.
\newblock \bibinfo{journal}{\emph{Information Fusion}} (\bibinfo{year}{2024}), \bibinfo{pages}{102317}.
\newblock


\bibitem[Gao et~al\mbox{.}(2007)]%
        {gao2007study}
\bibfield{author}{\bibinfo{person}{Zhonghua Gao}, \bibinfo{person}{Zhenjie Chen}, \bibinfo{person}{Yongxue Liu}, {and} \bibinfo{person}{Kang Huang}.} \bibinfo{year}{2007}\natexlab{}.
\newblock \showarticletitle{Study on the complex network characteristics of urban road system based on GIS}. In \bibinfo{booktitle}{\emph{Geoinformatics 2007: Geospatial Information Technology and Applications}}, Vol.~\bibinfo{volume}{6754}. International Society for Optics and Photonics, \bibinfo{pages}{67540N}.
\newblock


\bibitem[Geng et~al\mbox{.}(2019)]%
        {geng2019spatiotemporal}
\bibfield{author}{\bibinfo{person}{Xu Geng}, \bibinfo{person}{Yaguang Li}, \bibinfo{person}{Leye Wang}, \bibinfo{person}{Lingyu Zhang}, \bibinfo{person}{Qiang Yang}, \bibinfo{person}{Jieping Ye}, {and} \bibinfo{person}{Yan Liu}.} \bibinfo{year}{2019}\natexlab{}.
\newblock \showarticletitle{Spatiotemporal multi-graph convolution network for ride-hailing demand forecasting}. In \bibinfo{booktitle}{\emph{Proceedings of the AAAI conference on artificial intelligence}}, Vol.~\bibinfo{volume}{33}. \bibinfo{pages}{3656--3663}.
\newblock


\bibitem[Gilmer et~al\mbox{.}(2017)]%
        {gilmer2017mpnn}
\bibfield{author}{\bibinfo{person}{Justin Gilmer}, \bibinfo{person}{Samuel~S Schoenholz}, \bibinfo{person}{Patrick~F Riley}, \bibinfo{person}{Oriol Vinyals}, {and} \bibinfo{person}{George~E Dahl}.} \bibinfo{year}{2017}\natexlab{}.
\newblock \showarticletitle{Neural message passing for quantum chemistry}. In \bibinfo{booktitle}{\emph{International conference on machine learning}}. PMLR, \bibinfo{pages}{1263--1272}.
\newblock


\bibitem[Guo et~al\mbox{.}(2019)]%
        {guo2019attention}
\bibfield{author}{\bibinfo{person}{Shengnan Guo}, \bibinfo{person}{Youfang Lin}, \bibinfo{person}{Ning Feng}, \bibinfo{person}{Chao Song}, {and} \bibinfo{person}{Huaiyu Wan}.} \bibinfo{year}{2019}\natexlab{}.
\newblock \showarticletitle{Attention based spatial-temporal graph convolutional networks for traffic flow forecasting}. In \bibinfo{booktitle}{\emph{Proceedings of the AAAI conference on artificial intelligence}}, Vol.~\bibinfo{volume}{33}. \bibinfo{pages}{922--929}.
\newblock


\bibitem[Hamilton et~al\mbox{.}(2017)]%
        {hamilton2017inductive}
\bibfield{author}{\bibinfo{person}{William~L Hamilton}, \bibinfo{person}{Rex Ying}, {and} \bibinfo{person}{Jure Leskovec}.} \bibinfo{year}{2017}\natexlab{}.
\newblock \showarticletitle{Inductive representation learning on large graphs}. In \bibinfo{booktitle}{\emph{Proceedings of the 31st International Conference on Neural Information Processing Systems}}. \bibinfo{pages}{1025--1035}.
\newblock


\bibitem[Henaff et~al\mbox{.}(2015)]%
        {henaff2015deep}
\bibfield{author}{\bibinfo{person}{Mikael Henaff}, \bibinfo{person}{Joan Bruna}, {and} \bibinfo{person}{Yann LeCun}.} \bibinfo{year}{2015}\natexlab{}.
\newblock \showarticletitle{Deep convolutional networks on graph-structured data}.
\newblock \bibinfo{journal}{\emph{arXiv preprint arXiv:1506.05163}} (\bibinfo{year}{2015}).
\newblock


\bibitem[Huang et~al\mbox{.}(2023)]%
        {huang2023tap}
\bibfield{author}{\bibinfo{person}{Baixiang Huang}, \bibinfo{person}{Bryan Hooi}, {and} \bibinfo{person}{Kai Shu}.} \bibinfo{year}{2023}\natexlab{}.
\newblock \showarticletitle{TAP: A Comprehensive Data Repository for Traffic Accident Prediction in Road Networks}. In \bibinfo{booktitle}{\emph{Proceedings of the 31st ACM International Conference on Advances in Geographic Information Systems}}. \bibinfo{pages}{1--4}.
\newblock


\bibitem[Huang et~al\mbox{.}(2022)]%
        {huang2022gan}
\bibfield{author}{\bibinfo{person}{Ziheng Huang}, \bibinfo{person}{Weihan Zhang}, \bibinfo{person}{Dujuan Wang}, {and} \bibinfo{person}{Yunqiang Yin}.} \bibinfo{year}{2022}\natexlab{}.
\newblock \showarticletitle{A GAN framework-based dynamic multi-graph convolutional network for origin--destination-based ride-hailing demand prediction}.
\newblock \bibinfo{journal}{\emph{Information Sciences}}  \bibinfo{volume}{601} (\bibinfo{year}{2022}), \bibinfo{pages}{129--146}.
\newblock


\bibitem[Jiang and Luo(2022)]%
        {jiang2022graph}
\bibfield{author}{\bibinfo{person}{Weiwei Jiang} {and} \bibinfo{person}{Jiayun Luo}.} \bibinfo{year}{2022}\natexlab{}.
\newblock \showarticletitle{Graph neural network for traffic forecasting: A survey}.
\newblock \bibinfo{journal}{\emph{Expert Systems with Applications}}  \bibinfo{volume}{207} (\bibinfo{year}{2022}), \bibinfo{pages}{117921}.
\newblock


\bibitem[Jing et~al\mbox{.}(2022)]%
        {jing2022agnet}
\bibfield{author}{\bibinfo{person}{Weipeng Jing}, \bibinfo{person}{Wenjun Zhang}, \bibinfo{person}{Linhui Li}, \bibinfo{person}{Donglin Di}, \bibinfo{person}{Guangsheng Chen}, {and} \bibinfo{person}{Jian Wang}.} \bibinfo{year}{2022}\natexlab{}.
\newblock \showarticletitle{AGNet: An attention-based graph network for point cloud classification and segmentation}.
\newblock \bibinfo{journal}{\emph{Remote Sensing}} \bibinfo{volume}{14}, \bibinfo{number}{4} (\bibinfo{year}{2022}), \bibinfo{pages}{1036}.
\newblock


\bibitem[Ke et~al\mbox{.}(2021)]%
        {ke2021predicting}
\bibfield{author}{\bibinfo{person}{Jintao Ke}, \bibinfo{person}{Xiaoran Qin}, \bibinfo{person}{Hai Yang}, \bibinfo{person}{Zhengfei Zheng}, \bibinfo{person}{Zheng Zhu}, {and} \bibinfo{person}{Jieping Ye}.} \bibinfo{year}{2021}\natexlab{}.
\newblock \showarticletitle{Predicting origin-destination ride-sourcing demand with a spatio-temporal encoder-decoder residual multi-graph convolutional network}.
\newblock \bibinfo{journal}{\emph{Transportation Research Part C: Emerging Technologies}}  \bibinfo{volume}{122} (\bibinfo{year}{2021}), \bibinfo{pages}{102858}.
\newblock


\bibitem[Kipf and Welling(2016)]%
        {kipf2016gcnconv}
\bibfield{author}{\bibinfo{person}{Thomas~N Kipf} {and} \bibinfo{person}{Max Welling}.} \bibinfo{year}{2016}\natexlab{}.
\newblock \showarticletitle{Semi-supervised classification with graph convolutional networks}.
\newblock \bibinfo{journal}{\emph{arXiv preprint arXiv:1609.02907}} (\bibinfo{year}{2016}).
\newblock


\bibitem[Klinkhamer et~al\mbox{.}(2017)]%
        {klinkhamer2017functionally}
\bibfield{author}{\bibinfo{person}{Christopher Klinkhamer}, \bibinfo{person}{Elisabeth Krueger}, \bibinfo{person}{Xianyuan Zhan}, \bibinfo{person}{Frank Blumensaat}, \bibinfo{person}{Satish Ukkusuri}, {and} \bibinfo{person}{P~Suresh~C Rao}.} \bibinfo{year}{2017}\natexlab{}.
\newblock \showarticletitle{Functionally fractal urban networks: Geospatial co-location and homogeneity of infrastructure}.
\newblock \bibinfo{journal}{\emph{arXiv preprint arXiv:1712.03883}} (\bibinfo{year}{2017}).
\newblock


\bibitem[Li et~al\mbox{.}(2022)]%
        {li2022data}
\bibfield{author}{\bibinfo{person}{Guanyao Li}, \bibinfo{person}{Xiaofeng Wang}, \bibinfo{person}{Gunarto~Sindoro Njoo}, \bibinfo{person}{Shuhan Zhong}, \bibinfo{person}{S-H~Gary Chan}, \bibinfo{person}{Chih-Chieh Hung}, {and} \bibinfo{person}{Wen-Chih Peng}.} \bibinfo{year}{2022}\natexlab{}.
\newblock \showarticletitle{A data-driven spatial-temporal graph neural network for docked bike prediction}. In \bibinfo{booktitle}{\emph{2022 IEEE 38th International Conference on Data Engineering (ICDE)}}. IEEE, \bibinfo{pages}{713--726}.
\newblock


\bibitem[Li et~al\mbox{.}(2020)]%
        {li2020gen}
\bibfield{author}{\bibinfo{person}{Guohao Li}, \bibinfo{person}{Chenxin Xiong}, \bibinfo{person}{Ali Thabet}, {and} \bibinfo{person}{Bernard Ghanem}.} \bibinfo{year}{2020}\natexlab{}.
\newblock \bibinfo{title}{DeeperGCN: All You Need to Train Deeper GCNs}.
\newblock
\newblock
\showeprint[arxiv]{2006.07739}~[cs.LG]


\bibitem[Li et~al\mbox{.}(2017)]%
        {li2017speedrandomwalkgru}
\bibfield{author}{\bibinfo{person}{Yaguang Li}, \bibinfo{person}{Rose Yu}, \bibinfo{person}{Cyrus Shahabi}, {and} \bibinfo{person}{Yan Liu}.} \bibinfo{year}{2017}\natexlab{}.
\newblock \showarticletitle{Diffusion convolutional recurrent neural network: Data-driven traffic forecasting}.
\newblock \bibinfo{journal}{\emph{arXiv preprint arXiv:1707.01926}} (\bibinfo{year}{2017}).
\newblock


\bibitem[Lin et~al\mbox{.}(2018)]%
        {lin2018predicting}
\bibfield{author}{\bibinfo{person}{Lei Lin}, \bibinfo{person}{Zhengbing He}, {and} \bibinfo{person}{Srinivas Peeta}.} \bibinfo{year}{2018}\natexlab{}.
\newblock \showarticletitle{Predicting station-level hourly demand in a large-scale bike-sharing network: A graph convolutional neural network approach}.
\newblock \bibinfo{journal}{\emph{Transportation Research Part C: Emerging Technologies}}  \bibinfo{volume}{97} (\bibinfo{year}{2018}), \bibinfo{pages}{258--276}.
\newblock


\bibitem[Liu et~al\mbox{.}(2020)]%
        {liu2020physical}
\bibfield{author}{\bibinfo{person}{Lingbo Liu}, \bibinfo{person}{Jingwen Chen}, \bibinfo{person}{Hefeng Wu}, \bibinfo{person}{Jiajie Zhen}, \bibinfo{person}{Guanbin Li}, {and} \bibinfo{person}{Liang Lin}.} \bibinfo{year}{2020}\natexlab{}.
\newblock \showarticletitle{Physical-virtual collaboration modeling for intra-and inter-station metro ridership prediction}.
\newblock \bibinfo{journal}{\emph{IEEE Transactions on Intelligent Transportation Systems}} \bibinfo{volume}{23}, \bibinfo{number}{4} (\bibinfo{year}{2020}), \bibinfo{pages}{3377--3391}.
\newblock


\bibitem[Liu et~al\mbox{.}(2022)]%
        {liu2022contrastive}
\bibfield{author}{\bibinfo{person}{Xu Liu}, \bibinfo{person}{Yuxuan Liang}, \bibinfo{person}{Chao Huang}, \bibinfo{person}{Yu Zheng}, \bibinfo{person}{Bryan Hooi}, {and} \bibinfo{person}{Roger Zimmermann}.} \bibinfo{year}{2022}\natexlab{}.
\newblock \showarticletitle{When do contrastive learning signals help spatio-temporal graph forecasting?}. In \bibinfo{booktitle}{\emph{Proceedings of the 30th International Conference on Advances in Geographic Information Systems}}. \bibinfo{pages}{1--12}.
\newblock


\bibitem[{Microsoft Bing,}(2021)]%
        {Bing}
\bibfield{author}{\bibinfo{person}{{Microsoft Bing,}}.} \bibinfo{year}{2021}\natexlab{}.
\newblock \bibinfo{title}{Bing Map Traffic API}.
\newblock
\newblock
\urldef\tempurl%
\url{hhttps://www.bingmapsportal.com}
\showURL{%
\tempurl}


\bibitem[{Moosavi Sobhan}(2022)]%
        {US_accident}
\bibfield{author}{\bibinfo{person}{{Moosavi Sobhan}}.} \bibinfo{year}{2022}\natexlab{}.
\newblock \bibinfo{title}{US Accidents}.
\newblock
\newblock
\urldef\tempurl%
\url{https://www.kaggle.com/datasets/sobhanmoosavi/us-accidents}
\showURL{%
\tempurl}


\bibitem[Najjar et~al\mbox{.}(2017)]%
        {najjar2017imgaccident}
\bibfield{author}{\bibinfo{person}{Alameen Najjar}, \bibinfo{person}{Shun’ichi Kaneko}, {and} \bibinfo{person}{Yoshikazu Miyanaga}.} \bibinfo{year}{2017}\natexlab{}.
\newblock \showarticletitle{Combining satellite imagery and open data to map road safety}. In \bibinfo{booktitle}{\emph{Thirty-First AAAI Conference on Artificial Intelligence}}.
\newblock


\bibitem[{National Highway Traffic Safety Administration}(2023)]%
        {NHTSA}
\bibfield{author}{\bibinfo{person}{{National Highway Traffic Safety Administration}}.} \bibinfo{year}{2023}\natexlab{}.
\newblock \bibinfo{title}{NHTSA Report}.
\newblock
\newblock
\urldef\tempurl%
\url{https://crashstats.nhtsa.dot.gov/Api/Public/ViewPublication/813514}
\showURL{%
\tempurl}


\bibitem[Oh et~al\mbox{.}(2006)]%
        {oh2006regressionaccident}
\bibfield{author}{\bibinfo{person}{Jutaek Oh}, \bibinfo{person}{Simon~P Washington}, {and} \bibinfo{person}{Doohee Nam}.} \bibinfo{year}{2006}\natexlab{}.
\newblock \showarticletitle{Accident prediction model for railway-highway interfaces}.
\newblock \bibinfo{journal}{\emph{Accident Analysis \& Prevention}} \bibinfo{volume}{38}, \bibinfo{number}{2} (\bibinfo{year}{2006}), \bibinfo{pages}{346--356}.
\newblock


\bibitem[Pei and Hou(2024)]%
        {pei2024safety}
\bibfield{author}{\bibinfo{person}{Yulong Pei} {and} \bibinfo{person}{Lin Hou}.} \bibinfo{year}{2024}\natexlab{}.
\newblock \showarticletitle{Safety Assessment and Risk Management of Urban Arterial Traffic Flow Based on Artificial Driving and Intelligent Network Connection: An Overview}.
\newblock \bibinfo{journal}{\emph{Archives of Computational Methods in Engineering}} (\bibinfo{year}{2024}), \bibinfo{pages}{1--19}.
\newblock


\bibitem[Persaud and Dzbik(1992)]%
        {persaud1992accident}
\bibfield{author}{\bibinfo{person}{Bhagwant Persaud} {and} \bibinfo{person}{Leszek Dzbik}.} \bibinfo{year}{1992}\natexlab{}.
\newblock \showarticletitle{Accident prediction models for freeways}.
\newblock \bibinfo{journal}{\emph{Transportation Research Record}}  \bibinfo{volume}{1401} (\bibinfo{year}{1992}), \bibinfo{pages}{55--60}.
\newblock


\bibitem[Prince(2023)]%
        {prince2023understanding}
\bibfield{author}{\bibinfo{person}{Simon~J.D. Prince}.} \bibinfo{year}{2023}\natexlab{}.
\newblock \bibinfo{booktitle}{\emph{Understanding Deep Learning}}.
\newblock \bibinfo{publisher}{The MIT Press}.
\newblock
\urldef\tempurl%
\url{http://udlbook.com}
\showURL{%
\tempurl}


\bibitem[Rahmani et~al\mbox{.}(2023)]%
        {rahmani2023graph}
\bibfield{author}{\bibinfo{person}{Saeed Rahmani}, \bibinfo{person}{Asiye Baghbani}, \bibinfo{person}{Nizar Bouguila}, {and} \bibinfo{person}{Zachary Patterson}.} \bibinfo{year}{2023}\natexlab{}.
\newblock \showarticletitle{Graph neural networks for intelligent transportation systems: A survey}.
\newblock \bibinfo{journal}{\emph{IEEE Transactions on Intelligent Transportation Systems}} (\bibinfo{year}{2023}).
\newblock


\bibitem[Razi et~al\mbox{.}(2023)]%
        {razi2023deep}
\bibfield{author}{\bibinfo{person}{Abolfazl Razi}, \bibinfo{person}{Xiwen Chen}, \bibinfo{person}{Huayu Li}, \bibinfo{person}{Hao Wang}, \bibinfo{person}{Brendan Russo}, \bibinfo{person}{Yan Chen}, {and} \bibinfo{person}{Hongbin Yu}.} \bibinfo{year}{2023}\natexlab{}.
\newblock \showarticletitle{Deep learning serves traffic safety analysis: A forward-looking review}.
\newblock \bibinfo{journal}{\emph{IET Intelligent Transport Systems}} \bibinfo{volume}{17}, \bibinfo{number}{1} (\bibinfo{year}{2023}), \bibinfo{pages}{22--71}.
\newblock


\bibitem[Sarlak et~al\mbox{.}(2023)]%
        {sarlak2023diversity}
\bibfield{author}{\bibinfo{person}{Ahmad Sarlak}, \bibinfo{person}{Abolfazl Razi}, \bibinfo{person}{Xiwen Chen}, {and} \bibinfo{person}{Rahul Amin}.} \bibinfo{year}{2023}\natexlab{}.
\newblock \showarticletitle{Diversity maximized scheduling in roadside units for traffic monitoring applications}. In \bibinfo{booktitle}{\emph{2023 IEEE 48th Conference on Local Computer Networks (LCN)}}. IEEE, \bibinfo{pages}{1--4}.
\newblock


\bibitem[Scarselli et~al\mbox{.}(2008)]%
        {scarselli2008graph}
\bibfield{author}{\bibinfo{person}{Franco Scarselli}, \bibinfo{person}{Marco Gori}, \bibinfo{person}{Ah~Chung Tsoi}, \bibinfo{person}{Markus Hagenbuchner}, {and} \bibinfo{person}{Gabriele Monfardini}.} \bibinfo{year}{2008}\natexlab{}.
\newblock \showarticletitle{The graph neural network model}.
\newblock \bibinfo{journal}{\emph{IEEE transactions on neural networks}} \bibinfo{volume}{20}, \bibinfo{number}{1} (\bibinfo{year}{2008}), \bibinfo{pages}{61--80}.
\newblock


\bibitem[Shi and Rajkumar(2020)]%
        {shi2020point}
\bibfield{author}{\bibinfo{person}{Weijing Shi} {and} \bibinfo{person}{Raj Rajkumar}.} \bibinfo{year}{2020}\natexlab{}.
\newblock \showarticletitle{Point-gnn: Graph neural network for 3d object detection in a point cloud}. In \bibinfo{booktitle}{\emph{Proceedings of the IEEE/CVF conference on computer vision and pattern recognition}}. \bibinfo{pages}{1711--1719}.
\newblock


\bibitem[Shi et~al\mbox{.}(2021)]%
        {shi2021graphtransformer}
\bibfield{author}{\bibinfo{person}{Yunsheng Shi}, \bibinfo{person}{Zhengjie Huang}, \bibinfo{person}{Shikun Feng}, \bibinfo{person}{Hui Zhong}, \bibinfo{person}{Wenjin Wang}, {and} \bibinfo{person}{Yu Sun}.} \bibinfo{year}{2021}\natexlab{}.
\newblock \bibinfo{title}{Masked Label Prediction: Unified Message Passing Model for Semi-Supervised Classification}.
\newblock
\newblock
\showeprint[arxiv]{2009.03509}~[cs.LG]


\bibitem[Shin and Yoon(2020)]%
        {shin2020incorporating}
\bibfield{author}{\bibinfo{person}{Yuyol Shin} {and} \bibinfo{person}{Yoonjin Yoon}.} \bibinfo{year}{2020}\natexlab{}.
\newblock \showarticletitle{Incorporating dynamicity of transportation network with multi-weight traffic graph convolutional network for traffic forecasting}.
\newblock \bibinfo{journal}{\emph{IEEE Transactions on Intelligent Transportation Systems}} \bibinfo{volume}{23}, \bibinfo{number}{3} (\bibinfo{year}{2020}), \bibinfo{pages}{2082--2092}.
\newblock


\bibitem[Tang et~al\mbox{.}(2023)]%
        {tang2023trajectory}
\bibfield{author}{\bibinfo{person}{Luqi Tang}, \bibinfo{person}{Fuwu Yan}, \bibinfo{person}{Bin Zou}, \bibinfo{person}{Wenbo Li}, \bibinfo{person}{Chen Lv}, {and} \bibinfo{person}{Kewei Wang}.} \bibinfo{year}{2023}\natexlab{}.
\newblock \showarticletitle{Trajectory prediction for autonomous driving based on multiscale spatial-temporal graph}.
\newblock \bibinfo{journal}{\emph{IET Intelligent Transport Systems}} \bibinfo{volume}{17}, \bibinfo{number}{2} (\bibinfo{year}{2023}), \bibinfo{pages}{386--399}.
\newblock


\bibitem[Veličković et~al\mbox{.}(2018)]%
        {velivckovic2017gat}
\bibfield{author}{\bibinfo{person}{Petar Veličković}, \bibinfo{person}{Guillem Cucurull}, \bibinfo{person}{Arantxa Casanova}, \bibinfo{person}{Adriana Romero}, \bibinfo{person}{Pietro Liò}, {and} \bibinfo{person}{Yoshua Bengio}.} \bibinfo{year}{2018}\natexlab{}.
\newblock \showarticletitle{Graph Attention Networks}. In \bibinfo{booktitle}{\emph{International Conference on Learning Representations}}.
\newblock
\urldef\tempurl%
\url{https://openreview.net/forum?id=rJXMpikCZ}
\showURL{%
\tempurl}


\bibitem[Watts and Strogatz(1998)]%
        {watts1998collective}
\bibfield{author}{\bibinfo{person}{Duncan~J Watts} {and} \bibinfo{person}{Steven~H Strogatz}.} \bibinfo{year}{1998}\natexlab{}.
\newblock \showarticletitle{Collective dynamics of ‘small-world’networks}.
\newblock \bibinfo{journal}{\emph{nature}} \bibinfo{volume}{393}, \bibinfo{number}{6684} (\bibinfo{year}{1998}), \bibinfo{pages}{440--442}.
\newblock


\bibitem[Weisfeiler and Leman(1968)]%
        {weisfeiler1968reduction}
\bibfield{author}{\bibinfo{person}{Boris Weisfeiler} {and} \bibinfo{person}{Andrei Leman}.} \bibinfo{year}{1968}\natexlab{}.
\newblock \showarticletitle{The reduction of a graph to canonical form and the algebra which appears therein}.
\newblock \bibinfo{journal}{\emph{NTI, Series}} \bibinfo{volume}{2}, \bibinfo{number}{9} (\bibinfo{year}{1968}), \bibinfo{pages}{12--16}.
\newblock


\bibitem[Wu et~al\mbox{.}(2004)]%
        {wu2004urban}
\bibfield{author}{\bibinfo{person}{Jianjun Wu}, \bibinfo{person}{Ziyou Gao}, \bibinfo{person}{Huijun Sun}, {and} \bibinfo{person}{Haijun Huang}.} \bibinfo{year}{2004}\natexlab{}.
\newblock \showarticletitle{Urban transit system as a scale-free network}.
\newblock \bibinfo{journal}{\emph{Modern Physics Letters B}} \bibinfo{volume}{18}, \bibinfo{number}{19n20} (\bibinfo{year}{2004}), \bibinfo{pages}{1043--1049}.
\newblock


\bibitem[Xie and Grossman(2018)]%
        {xie2018cgc}
\bibfield{author}{\bibinfo{person}{Tian Xie} {and} \bibinfo{person}{Jeffrey~C. Grossman}.} \bibinfo{year}{2018}\natexlab{}.
\newblock \showarticletitle{Crystal Graph Convolutional Neural Networks for an Accurate and Interpretable Prediction of Material Properties}.
\newblock \bibinfo{journal}{\emph{Phys. Rev. Lett.}}  \bibinfo{volume}{120} (\bibinfo{date}{Apr} \bibinfo{year}{2018}), \bibinfo{pages}{145301}.
\newblock
Issue 14.
\urldef\tempurl%
\url{https://doi.org/10.1103/PhysRevLett.120.145301}
\showDOI{\tempurl}


\bibitem[Xu et~al\mbox{.}(2018)]%
        {xu2018gnn}
\bibfield{author}{\bibinfo{person}{Keyulu Xu}, \bibinfo{person}{Weihua Hu}, \bibinfo{person}{Jure Leskovec}, {and} \bibinfo{person}{Stefanie Jegelka}.} \bibinfo{year}{2018}\natexlab{}.
\newblock \showarticletitle{How powerful are graph neural networks?}
\newblock \bibinfo{journal}{\emph{arXiv preprint arXiv:1810.00826}} (\bibinfo{year}{2018}).
\newblock


\bibitem[Ye et~al\mbox{.}(2020)]%
        {ye2020build}
\bibfield{author}{\bibinfo{person}{Jiexia Ye}, \bibinfo{person}{Juanjuan Zhao}, \bibinfo{person}{Kejiang Ye}, {and} \bibinfo{person}{Chengzhong Xu}.} \bibinfo{year}{2020}\natexlab{}.
\newblock \showarticletitle{How to build a graph-based deep learning architecture in traffic domain: A survey}.
\newblock \bibinfo{journal}{\emph{IEEE Transactions on Intelligent Transportation Systems}} \bibinfo{volume}{23}, \bibinfo{number}{5} (\bibinfo{year}{2020}), \bibinfo{pages}{3904--3924}.
\newblock


\bibitem[Yin et~al\mbox{.}(2021)]%
        {yin2021deep}
\bibfield{author}{\bibinfo{person}{Xueyan Yin}, \bibinfo{person}{Genze Wu}, \bibinfo{person}{Jinze Wei}, \bibinfo{person}{Yanming Shen}, \bibinfo{person}{Heng Qi}, {and} \bibinfo{person}{Baocai Yin}.} \bibinfo{year}{2021}\natexlab{}.
\newblock \showarticletitle{Deep learning on traffic prediction: Methods, analysis, and future directions}.
\newblock \bibinfo{journal}{\emph{IEEE Transactions on Intelligent Transportation Systems}} \bibinfo{volume}{23}, \bibinfo{number}{6} (\bibinfo{year}{2021}), \bibinfo{pages}{4927--4943}.
\newblock


\bibitem[You et~al\mbox{.}(2020)]%
        {you2020graph}
\bibfield{author}{\bibinfo{person}{Yuning You}, \bibinfo{person}{Tianlong Chen}, \bibinfo{person}{Yongduo Sui}, \bibinfo{person}{Ting Chen}, \bibinfo{person}{Zhangyang Wang}, {and} \bibinfo{person}{Yang Shen}.} \bibinfo{year}{2020}\natexlab{}.
\newblock \showarticletitle{Graph contrastive learning with augmentations}.
\newblock \bibinfo{journal}{\emph{Advances in neural information processing systems}}  \bibinfo{volume}{33} (\bibinfo{year}{2020}), \bibinfo{pages}{5812--5823}.
\newblock


\bibitem[Yu et~al\mbox{.}(2021)]%
        {yu2021deep}
\bibfield{author}{\bibinfo{person}{Le Yu}, \bibinfo{person}{Bowen Du}, \bibinfo{person}{Xiao Hu}, \bibinfo{person}{Leilei Sun}, \bibinfo{person}{Liangzhe Han}, {and} \bibinfo{person}{Weifeng Lv}.} \bibinfo{year}{2021}\natexlab{}.
\newblock \showarticletitle{Deep spatio-temporal graph convolutional network for traffic accident prediction}.
\newblock \bibinfo{journal}{\emph{Neurocomputing}}  \bibinfo{volume}{423} (\bibinfo{year}{2021}), \bibinfo{pages}{135--147}.
\newblock


\bibitem[Zhang et~al\mbox{.}(2024)]%
        {zhang2024beyond}
\bibfield{author}{\bibinfo{person}{Bohang Zhang}, \bibinfo{person}{Jingchu Gai}, \bibinfo{person}{Yiheng Du}, \bibinfo{person}{Qiwei Ye}, \bibinfo{person}{Di He}, {and} \bibinfo{person}{Liwei Wang}.} \bibinfo{year}{2024}\natexlab{}.
\newblock \showarticletitle{Beyond Weisfeiler-Lehman: A Quantitative Framework for {GNN} Expressiveness}. In \bibinfo{booktitle}{\emph{The Twelfth International Conference on Learning Representations}}.
\newblock
\urldef\tempurl%
\url{https://openreview.net/forum?id=HSKaGOi7Ar}
\showURL{%
\tempurl}


\bibitem[Zhang et~al\mbox{.}(2023)]%
        {zhang2023multiclass}
\bibfield{author}{\bibinfo{person}{Xinan Zhang}, \bibinfo{person}{Yung-An Hsieh}, \bibinfo{person}{Pingzhou Yu}, \bibinfo{person}{Zhongyu Yang}, {and} \bibinfo{person}{Yichang~James Tsai}.} \bibinfo{year}{2023}\natexlab{}.
\newblock \showarticletitle{Multiclass Transportation Safety Hardware Asset Detection and Segmentation Based on Mask-RCNN with RoI Attention and IoMA-Merging}.
\newblock \bibinfo{journal}{\emph{Journal of Computing in Civil Engineering}} \bibinfo{volume}{37}, \bibinfo{number}{5} (\bibinfo{year}{2023}), \bibinfo{pages}{04023024}.
\newblock


\bibitem[Zhao et~al\mbox{.}(2019)]%
        {zhao2019t}
\bibfield{author}{\bibinfo{person}{Ling Zhao}, \bibinfo{person}{Yujiao Song}, \bibinfo{person}{Chao Zhang}, \bibinfo{person}{Yu Liu}, \bibinfo{person}{Pu Wang}, \bibinfo{person}{Tao Lin}, \bibinfo{person}{Min Deng}, {and} \bibinfo{person}{Haifeng Li}.} \bibinfo{year}{2019}\natexlab{}.
\newblock \showarticletitle{T-gcn: A temporal graph convolutional network for traffic prediction}.
\newblock \bibinfo{journal}{\emph{IEEE transactions on intelligent transportation systems}} \bibinfo{volume}{21}, \bibinfo{number}{9} (\bibinfo{year}{2019}), \bibinfo{pages}{3848--3858}.
\newblock


\bibitem[Zheng et~al\mbox{.}(2021)]%
        {zheng2021modeling}
\bibfield{author}{\bibinfo{person}{Lai Zheng}, \bibinfo{person}{Tarek Sayed}, {and} \bibinfo{person}{Fred Mannering}.} \bibinfo{year}{2021}\natexlab{}.
\newblock \showarticletitle{Modeling traffic conflicts for use in road safety analysis: A review of analytic methods and future directions}.
\newblock \bibinfo{journal}{\emph{Analytic methods in accident research}}  \bibinfo{volume}{29} (\bibinfo{year}{2021}), \bibinfo{pages}{100142}.
\newblock


\bibitem[Zhou et~al\mbox{.}(2020a)]%
        {zhou2020riskoracle}
\bibfield{author}{\bibinfo{person}{Zhengyang Zhou}, \bibinfo{person}{Yang Wang}, \bibinfo{person}{Xike Xie}, \bibinfo{person}{Lianliang Chen}, {and} \bibinfo{person}{Hengchang Liu}.} \bibinfo{year}{2020}\natexlab{a}.
\newblock \showarticletitle{RiskOracle: A Minute-Level Citywide Traffic Accident Forecasting Framework}. In \bibinfo{booktitle}{\emph{Proceedings of the AAAI Conference on Artificial Intelligence}}, Vol.~\bibinfo{volume}{34}. \bibinfo{pages}{1258--1265}.
\newblock


\bibitem[Zhou et~al\mbox{.}(2020b)]%
        {zhou2020foresee}
\bibfield{author}{\bibinfo{person}{Zhengyang Zhou}, \bibinfo{person}{Yang Wang}, \bibinfo{person}{Xike Xie}, \bibinfo{person}{Lianliang Chen}, {and} \bibinfo{person}{Chaochao Zhu}.} \bibinfo{year}{2020}\natexlab{b}.
\newblock \showarticletitle{Foresee urban sparse traffic accidents: A spatiotemporal multi-granularity perspective}.
\newblock \bibinfo{journal}{\emph{IEEE Transactions on Knowledge and Data Engineering}} \bibinfo{volume}{34}, \bibinfo{number}{8} (\bibinfo{year}{2020}), \bibinfo{pages}{3786--3799}.
\newblock


\end{thebibliography}




\end{document}